\documentclass{article}

\PassOptionsToPackage{numbers, compress}{natbib}
\usepackage[main,final]{neurips_2026}

\usepackage[utf8]{inputenc} 
\usepackage[T1]{fontenc}    
\usepackage{hyperref}       
\usepackage{url}            
\usepackage{booktabs}       
\usepackage{amsfonts}       
\usepackage{nicefrac}       
\usepackage{xcolor}         
\usepackage{bm}
\usepackage{amsmath}
\usepackage{wrapfig}
\usepackage{graphicx}
\usepackage{comment}
\usepackage{tabularx}
\usepackage{booktabs}
\usepackage{subcaption}
\usepackage{makecell}
\usepackage{multirow}
\usepackage{float}
\usepackage{enumitem}
\usepackage{placeins}
\usepackage{makecell}
\usepackage{url}

\title{Symb-xMIL: Symbolic Explanations for Multiple Instance Learning in Digital Pathology}

\author{%
    \textbf{Yanqing Luo}$^{1,2}$ \quad
    \textbf{Julius Hense}$^{1,2}$ \quad
    \textbf{Niklas Prenißl}$^{3,4}$ \\
    \textbf{Andreas Mock}$^{5,6,7}$ \quad
    \textbf{Klaus-Robert Müller}$^{1,2,8,9}$ \\
    \textbf{Thomas Schnake}$^{10,11,12,\dagger}$ \quad
    \textbf{Mina Jamshidi Idaji}$^{1,2,\dagger}$ \vspace{2mm} \vspace{2mm} \\
    $^{1}$Berlin Institute for the Foundations of Learning and Data (BIFOLD), Berlin, Germany \\
    $^{2}$Machine Learning Group, Technische Universität Berlin, Berlin, Germany \\
    $^{3}$Institute of Pathology, Charité Universitätsmedizin, Berlin, Germany  \\
    $^{4}$Berlin Institute of Health at Charité -- Universitätsmedizin Berlin, \\BIH Biomedical Innovation Academy, BIH Charité Digital Clinician Scientist Program, Berlin, Germany\\
    $^{5}$Institute of Pathology, Ludwig Maximilian University of Munich, Munich, Germany \\
    $^{6}$Division of Translational Medical Oncology, DKFZ, Heidelberg, Germany; \\ NCT Heidelberg, Heidelberg, Germany \\
    $^{7}$German Cancer Consortium (DKTK), partner site Munich, \\a partnership between DKFZ and Ludwig-Maximilians-Universität München (LMU), Germany \\
    $^{8}$Department of Artificial Intelligence, Korea University, Seoul, Korea \\
    $^{9}$Max-Planck Institute for Informatics, Saarbrücken, Germany \\
    $^{10}$ Department of Chemistry, Chemical Physics Theory Group, University of Toronto, Canada \\
    $^{11}$ Vector Institute for Artificial Intelligence, Toronto, Canada \\
    $^{12}$ Acceleration Consortium, University of Toronto, Canada \\
    $^{\dagger}$Equal supervision contribution.\\
    $^{\dagger}$ mina.jamshidi.idaji@tu-berlin.de, t.schnake@utoronto.ca
}

\begin{document}

\maketitle

\begin{abstract}
Explanations of multiple instance learning (MIL) models are widely used for validation and discovery in digital histopathology. Existing methods primarily rely on heatmaps that highlight influential regions but do not explain \textit{how} evidence from different tissue regions is combined to produce a prediction. This limits interpretability, especially when decisions depend on interactions between tissue features. We introduce Symbolic explainable MIL (Symb-xMIL), a post-hoc explanation framework that quantifies how a MIL model’s behavior aligns with human-readable decision rules, expressed as logical relationships (e.g., AND, OR, NOT) between input features. These alignment scores reveal semantic patterns underlying the model’s predictions. We evaluate Symb-xMIL on synthetic and real-world histopathology datasets. On synthetic MIL data, Symb-xMIL reliably recovers ground-truth logical rules. In a clinical tumor detection task, the best-aligned rules uncover heterogeneous decision patterns and expose hidden model errors. On an HPV-prediction task on TCGA-HNSCC, a cohort of head and neck cancer, our framework refines patient survival stratification beyond HPV status with potential clinical relevance. Overall, Symb-xMIL extends MIL explainability beyond visual attribution toward structured, rule-based reasoning, enabling more transparent and semantically grounded interpretation of model predictions. Code is available at \url{https://github.com/bifold-pathomics/Symb-xMIL}.

\end{abstract}

\section{Introduction}
Digital pathology enables large-scale analysis of gigapixel whole-slide images (WSIs) for diagnostic and prognostic tasks~\cite{campanella2019clinical, louis2016computational, koemen2025pathorob}. Despite the progress of deep learning models, their clinical adoption is still lacking, as transparent reasoning—among other factors—continues to be essential for medical decision-making~\cite{cui2021artificial,evans2022explainability}. Among various approaches, Multiple Instance Learning (MIL)~\cite{dietterich1997solving,ilse2018attention,shao2021transmil,cai2025attrimil} has become a popular paradigm for digital pathology tasks~\cite{klauschen2024toward, hagele2020resolving,lee2022derivation}, where each WSI is partitioned into patches that collectively form a bag of instances representing the slide. 

Current explainability methods \cite{samek2021xai_review} for MIL in digital pathology \cite{early2022model,hense2024xmil,idaji2026beyond,sundararajan2017axiomatic,binder2021morphological} predominantly rely on spatial heatmaps that highlight regions of a WSI attended to by the model, but suffer from several limitations. First, interpreting heatmaps requires substantial domain expertise to infer underlying morphological patterns; even then, the reasoning remains qualitative and difficult to compare across samples. Second, heatmap-based explanations do not readily support systematic analyses at the cohort level, where one aims to study how model behavior varies across cases and relates to predictions. Third, spatial heatmaps provide only \textit{first-order} explanations, quantifying the contribution of individual features, but fail to capture higher-order interactions, i.e., how combinations of features jointly influence predictions \cite{schnake2025towards,eberle2020building,schnake2021higher,xiong2022efficient,janizek2021explaining}. In histopathology, such interactions can reveal how the interplay between tumor and microenvironment shapes the model’s prediction.

To address these limitations, we augment heatmap-based explanations with a structured, symbolic representation of model reasoning, enabling both local and global interpretability \cite{armgaan2024graphtrail,guidotti2019factual,schnake2025towards}. Specifically, we evaluate how well the model’s response patterns align with a predefined set of logical rules. Based on this idea, we propose \textbf{Symbolic explainable MIL (Symb-xMIL)}, a framework that extends Symbolic XAI \cite{schnake2025towards} to MIL with applications in digital pathology.

Mapping model behavior to logical rules provides two key advantages. First, it enables the identification of the rule that best explains the model’s prediction for a given sample. Second, it transforms the model’s decision strategy into a vector space, where each dimension corresponds to a concrete, interpretable decision rule, yielding a high-dimensional \textit{symbolic representation space}. Unlike heatmaps, this representation is directly comparable across samples (e.g., patients in histopathology) and captures both the dominant rule and the full spectrum of competing explanations. 

This enables, for the first time, a systematic and quantitative analysis of model reasoning at the cohort level, going beyond what is possible with heatmap-based explanations. In particular, it allows samples to be grouped based on similarity in their semantic profiles, reveals distinct decision strategies within a cohort, and enables the systematic characterization of both typical and atypical reasoning patterns.

Our contributions are as follows:
\begin{enumerate}[leftmargin=9pt]
    \item \textbf{Symbolic explainable MIL (Symb-xMIL).} We introduce a post-hoc explanation framework for multiple instance learning that moves beyond first-order heatmaps toward symbolic, rule-based explanations. Symb-xMIL captures how \textit{combinations of features} jointly contribute to bag-level predictions, enabling interpretable reasoning at the slide level.

    \item \textbf{Symbolic representation space.} We propose a \textit{symbolic representation space} that embeds each sample according to its alignment with a set of logical rules. This representation provides a structured and comparable view of model behavior, linking predictions to interpretable reasoning patterns.

    \item \textbf{Empirical validation and insights.} We validate Symb-xMIL on synthetic and real-world histopathology datasets, showing that it recovers ground-truth rules and uncovers meaningful structure in model behavior. In particular, Symb-xMIL identifies distinct prediction strategies and reveals clinically relevant patient subgroups in TCGA-HNSCC.
\end{enumerate}

\section{Related work}\label{sec:related_work}

\textbf{Multiple Instance Learning (MIL).}
In MIL, a sample is represented as a bag of instances \( \mathbf{X} = \{\mathbf{x}_1, \ldots, \mathbf{x}_n\} \), and a model learns to aggregate instance-level information into a bag-level prediction \( f(\mathbf{X}) \approx y \)~\cite{dietterich1997solving}. Classical approaches rely on simple pooling mechanisms such as max or mean~\cite{dietterich1997solving,fu2012implementation,maron1997framework}, while modern methods employ attention-based pooling or transformers to model instance importance and inter-instance dependencies~\cite{gao2024transformer,ilse2018attention, shao2021transmil, shi2025positional,zhang2023multi}. Despite these advances, the resulting decision processes remain difficult to interpret, particularly as models capture increasingly complex interactions between instances.

\textbf{Explaining MIL.}
Existing explanation methods for MIL predominantly provide instance-wise relevance scores, visualized as spatial heatmaps~\cite{hense2024xmil,idaji2026beyond,ilse2018attention,sundararajan2017axiomatic}. While useful for localizing relevant instances, instance-wise heatmaps do not explicitly characterize how evidence is \emph{combined} across instances, such as whether features must co-occur, act as interchangeable signals, or interact in more complex ways. Prior work has begun to explore such interactions, for example by extracting pairwise relationships from attention mechanisms~\cite{jaume24dense}, but remains restricted to low-order interactions and specific model architectures. A general and interpretable characterization of multi-instance decision logic remains an open challenge. Building on Symbolic XAI~\cite{schnake2025towards} (see below), we address this challenge by characterizing model responses to subsets of instances through their alignment with logical queries. Unlike attribution methods, our alignment scores quantify agreement between the ML model and a given logical rule.

\textbf{Concept-based MIL and symbolic explanation.}
Recent approaches improve MIL interpretability through prototypes or semantic concepts. ProtoMIL~\cite{rymarczyk2022protomil} explains predictions via similarity to learned prototypes, and ConceptMIL~\cite{sun2025label} assigns semantic concepts to salient patches, and SI-MIL~\cite{kapse2024si} uses an interpretable branch grounded in handcrafted pathological features. These methods use the interpretable concepts during training and remain focused on individual prototypes or concepts without explicitly characterizing the relationships among them. Our goal is complementary: to characterize how an already-trained MIL model responds to different combinations of features without modifying or retraining it.
In parallel, Symbolic XAI provides a principled framework for aligning model behavior with logical rules, enabling interpretable descriptions of feature interactions. However, it is limited to standard supervised settings and does not directly apply to MIL, where predictions arise from aggregating sets of instances. Extending symbolic explanations to this setting requires modeling how subsets of instances jointly contribute to bag-level predictions,
enabling slide-level symbolic explanations over semantic instance concepts and a symbolic representation space for cohort-level analysis.

\section{Methodology}
\begin{figure*}[!t]
    \centering
    \includegraphics[width=\textwidth]{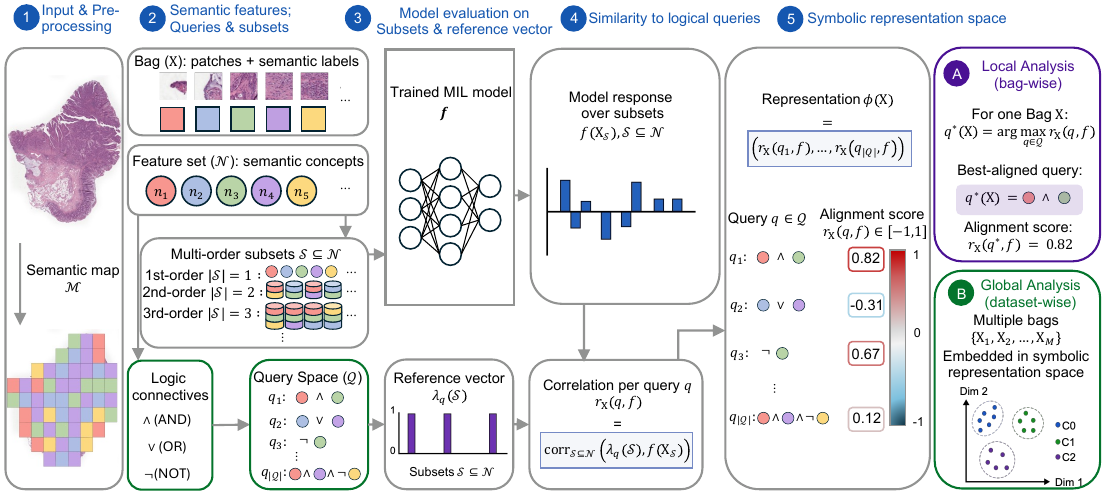}
    \caption{\textbf{Overview of the Symb-xMIL pipeline.} \textbf{(1) Input \& Pre-processing:} Patches are extracted from a WSI, and a predefined semantic map $\mathcal{M}$ assigns each patch a semantic label. \textbf{(2) Semantic features, queries \& subsets:} Patches are grouped by semantic features to construct sub-bags $X_S$ for feature subsets $S \subseteq N$. Logical queries combine features using predefined operators. \textbf{(3) Model evaluation on subsets and reference vector:} The MIL model is evaluated on each sub-bag ($\left[f(X_S)\right]_{S \subseteq N}$). Each logical query $q$ defines a corresponding reference vector $\left[\lambda_q(S)\right]_{S \subseteq N}$ over the same subsets. \textbf{(4) Similarity to logical queries:} For each slide and each logical query, we compute an alignment score using the Pearson correlation between the query's reference vector and the slide's model response vector. \textbf{(5) Symbolic representation space:} The scores form $\phi(X)$, with one dimension per query, supporting \textbf{(A) local analysis} through the best-aligned query for an individual sample and \textbf{(B) global analysis} through cohort-level comparisons such as clustering.}
    \label{fig:method}
    \vspace{-1em}
\end{figure*}

The Symbolic XAI framework \cite{schnake2025towards} explains machine learning models by aligning their behavior with logical relationships between input features. Given a logical formula $q$, the framework quantifies how well the model’s prediction aligns with this rule and defines a relevance measure for $q$ with respect to the prediction. In this work, we extend this framework by refining the alignment formulation and representing model behavior in a structured \textit{symbolic representation space}. We further generalize this approach to the MIL setting, resulting in \textit{Symbolic explainable MIL (Symb-xMIL)}.

\subsection{Background: Symbolic XAI}
\label{method:sumbolic_xai}
Formally, assume an ML model $f$ mapping feature vectors $\mathbf{x}=(x_I)_{I\in \mathcal{N}}$ to a scalar-valued prediction target, with $\mathcal{N} = \{1, \dots, n\}$ denoting the index set of all features. For any subset $\mathcal{S} \subseteq \mathcal{N}$, we define $\mathbf{x}_\mathcal{S}$ as a feature vector for which $(\mathbf{x}_\mathcal{S})_I = \mathbf{x}_I$ for $I\in \mathcal{S}$, while the remaining features are inpainted \cite{schnake2025towards}, i.e., replaced by a placeholder value.

The Symbolic XAI framework \cite{schnake2025towards} analyzes the model's prediction strategy by evaluating its behavior on $(\mathbf{x}_\mathcal{S})_{\mathcal{S} \subseteq \mathcal{N}}$, i.e., by systematically studying $f(\mathbf{x}_\mathcal{S})$. Alternative approaches within Symbolic XAI track model behavior on input subsets using, for example, propagation-based methods.
 
The framework further introduces the notion of \textit{queries}, which are relationships constructed by combining feature identifiers $I, J \in \mathcal{N}$ with logical connectives such as $\wedge$, $\vee$, and $\neg$, representing AND, OR, and NOT, respectively. For example, queries may take the form $q_1 = I \,\wedge\, J$ or $q_2 = I\, \vee\, \neg J$. We denote by $Q$ the set of all well-formed queries. Details of query generation in this work are included in Appendix~\ref{appendix:query_space}.   

Formally, a query $q$ is represented by a set function $\lambda_q : 2^\mathcal{N} \rightarrow \{0,1\}$ that maps each subset $\mathcal{S} \subseteq \mathcal{N}$ to a binary value. Specifically, $\lambda_q(\mathcal{S})$ indicates whether the logical condition defined by $q$ is satisfied within $\mathcal{S}$. For instance, for the queries $q_1$ and $q_2$ defined above, we have $\lambda_{q_1}(\mathcal{S}) = \bm{1}_{\{I \in \mathcal{S} \wedge J \in \mathcal{S}\}}$ and $\lambda_{q_2}(\mathcal{S}) = \bm{1}_{\{I \in \mathcal{S} \vee J \notin \mathcal{S}\}}$. 

Symbolic XAI quantifies the \textit{alignment} of an ML model $f$'s decision strategy with a specific query $q$ by comparing $\lambda_q$ with the Harsanyi dividend of $f(\mathbf{x}_\mathcal{S})$—a well-known transformation from cooperative game theory that is widely used in XAI.

\subsection{Symbolic explainable MIL (Symb-xMIL)}\label{sec:symbolic_MIL}
\label{method:symb_mil}
We argue that, for our purposes, the Harsanyi dividends of the model's prediction are not directly comparable to the non-transformed set function $\lambda_q$ of a query $q$. We therefore propose a modification that directly compares the model's prediction $f(\mathbf{x}_\mathcal{S})$ with $\lambda_q(\mathcal{S})$, without applying this transformation. In other words, we assume an \textit{idealized labeling function} that exactly follows the logical rule defined by $q$, and quantify its alignment with the model.

We measure similarity between a logical formula $q$ and the model by evaluating $f(\mathbf{x}_\mathcal{S})$ and $\lambda_q(\mathcal{S})$ across subsets $\mathcal{S} \subseteq \mathcal{N}$. If both functions exhibit similar behavior over these subsets, the formula $q$ provides a faithful representation of the model's behavior.

We quantify this alignment using correlation:
\begin{align}\label{eq:sybm_xmil_corr}
    r_{\mathbf{X}}(q, f) := 
    \operatorname{corr}_{\mathcal{S} \subseteq \mathcal{N}} \!\left(\lambda_q(\mathcal{S}),\, f(\mathbf{x}_\mathcal{S}) \right).
\end{align}
We refer to this as the \textit{alignment score} between $q$ and $f$ for $\mathbf{X}$ defined on $\mathcal{N}$. In Table~\ref{tab:logical_rules_interpretation}, we provide example queries $q$ and their corresponding interpretations when the model assigns a high value to $r_{\mathbf{X}}(q, f)$ in Eq.~\eqref{eq:sybm_xmil_corr}.

\begin{table}[t!]
\centering
\caption{\textbf{Natural language interpretations of common logical rules when their alignment to the model is high} (i.e., high \textit{alignment score}, cf.~Eq.~\eqref{eq:sybm_xmil_corr}). The formulas are defined over features $\texttt{A}$ and $\texttt{B}$, with $\mathcal{N}$ denoting the set of all feature indices.}

\begin{tabular}{l p{0.7\textwidth}}
\toprule
\textbf{Logical Formula $q$} & \textbf{Interpretation for high alignment score} \\
\midrule
$\texttt{A} \land \texttt{B}$ 
& The prediction is driven by the joint presence of $\texttt{A}$ and $\texttt{B}$. \\[0.5em]

$\texttt{A} \lor \texttt{B}$ 
& The prediction is supported by $\texttt{A}$, $\texttt{B}$, or both, indicating that each feature provides relevant evidence independently. \\[0.5em]

$\neg\texttt{A}$ ($=\neg\texttt{A} \land \texttt{Rest}$) 
& The prediction is driven by features other than $\texttt{A}$. Feature $\texttt{A}$ provides evidence against the prediction, while the remaining features support it. 
Here, $\texttt{Rest} = \mathcal{N} \setminus \{\texttt{A}\}$. \\[0.5em]

$\neg\texttt{A} \lor \texttt{B}$ ($=\texttt{A} \rightarrow \texttt{B}$) 
& The model predicts positively either in the absence of $\texttt{A}$, or—if $\texttt{A}$ is present—only when $\texttt{B}$ is also present. Thus, $\texttt{A}$ provides negative evidence unless compensated by $\texttt{B}$. \\
\bottomrule
\end{tabular}
\label{tab:logical_rules_interpretation}
\vspace{-0.8em}
\end{table}

Additionally, we extend this framework to the MIL setting. For a bag $\mathbf{X} = \{\mathbf{x}_1, \dots, \mathbf{x}_n\}$, we define a semantic map $\mathcal{M}$ that assigns each instance a semantic value. For example, in digital pathology, each patch of a whole slide image (WSI) can be assigned a tissue type. Such interpretable semantic categories are often referred to as \textit{semantic concepts} in the literature; here, we refer to them as \textit{features} in our framework. In this context, we define $\mathcal{N}$ as the set of all semantic values. Accordingly, $\mathbf{X}_{\mathcal{S}} = \{\mathbf{x}_i  \mid \mathcal{M}(\mathbf{x}_i) \in \mathcal{S}\}$ denotes the sub-bag containing all instances whose semantic values lie in $\mathcal{S}$. For instance, if $\mathbf{X}$ is a WSI and $\mathcal{S} = \{\texttt{Tumor}, \texttt{Stroma}\}$, then $\mathbf{X}_{\mathcal{S}}$ consists of all patches assigned to \texttt{Tumor} or \texttt{Stroma}. 

Symb-xMIL then uses predictions over such sub-bags, $f(\mathbf{X}_{\mathcal{S}})$ for $\mathcal{S} \subseteq \mathcal{N}$, to align the model’s decision strategy for a bag $\mathbf{X}$ with a logical rule represented by a query $q$, as defined in Eq.~\eqref{eq:sybm_xmil_corr}. For each bag, the best-aligned query is defined as 
\begin{align}\label{eq:q_star}
    q^*(\mathbf{X}) := \operatorname*{arg\,max}_{q \in Q}\, r_{\mathbf{X}}(q, f)
\end{align}

This formulation enables interpretable reasoning over how semantically defined groups of instances jointly contribute to bag-level predictions, which is not accessible with instance-level explanations.

\subsection{Symbolic representation space}
\label{method_symbolic_space}

For a given sample $\mathbf{X}$, we compute the alignment scores $\{r_{\mathbf{X}}(q, f)\}_{q \in Q}$. These scores define a vector-valued representation
\begin{align}\label{eq:symbolic_space}
    \phi(\mathbf{X}) = (r_{\mathbf{X}}(q_1, f), \dots, r_{\mathbf{X}}(q_{|Q|}, f))
\end{align}
where $|Q|$ is the number of queries in $Q$. $\phi(\mathbf{X})$ can be interpreted as a representation of the model’s behavior in a symbolic space induced by the query set $Q$. In this space, each dimension corresponds to alignment with a specific logical rule in $Q$. This representation projects the model’s reasoning onto a predefined set of queries. Details of query generation in this work, i.e., how $Q$ is constructed, are provided in Appendix~\ref{appendix:query_space}. 

The resulting \textit{symbolic representation space} maps each sample to a vector encoding the model’s reasoning, enabling direct and systematic comparison of model behavior across samples and forming the basis for both local and global downstream analyses. This formulation provides, for the first time, a structured representation of model reasoning at the cohort level.

\section{Experiments and results}
To demonstrate the value of Symb-xMIL, we verify in Section~\ref{experiment_conceptual} in a controlled simulation setting that Symb-xMIL can recover ground-truth logical rules, which cannot be captured by other methods. We then showcase the practical utility of the method in two real-world histopathology experiments. We show that Symb-xMIL reveals previously undetected model errors in a diagnostic setting in Section~\ref{experiment_camelyon16}. Besides, it exposes the prognostic value of heterogeneous model strategies for HPV-prediction in head-and-neck tumors in Section~\ref{experiment_hnsc}.

\subsection{Recovering higher-order feature interactions in  simulated settings}
\label{experiment_conceptual}

We designed simulated scenarios to validate whether Symb-xMIL explanations can accurately identify the symbolic decision rule in a MIL task. We constructed synthetic MNIST-based MIL datasets with binary targets \cite{early2022model,hense2024xmil, ilse2018attention}. Specifically, we drew bags of instances by randomly sampling images from MNIST~\cite{deng2012mnist}, with each instance $\mathbf{x}_i$ of a bag $\mathbf{X}=\{\mathbf{x}_i\}_i$ being associated through the semantic map $\mathcal{M}$ with a digit, i.e. $\mathcal{M}(\mathbf{x}_i) \in \{\texttt{0}, \texttt{1}, \cdots, \texttt{9}\} = \mathcal{N}$. We then define the bag label $y$ using three logical rules based on the operators $\land$ (AND), $\lor$ (OR), and $\neg$ (NOT) (see Table~\ref{tab:mnist_rules}).

\begin{table}[t!]
\vspace{-0.8em}
\centering
\caption{\textbf{Logical rules used to define bag labels in the synthetic MIL experiments.} Here, $\texttt{k} \in \mathcal{M}(\mathbf{X})$ denotes that at least one instance in the bag has digit label $k$, where $\mathcal{M}(\mathbf{X}) = \{ \mathcal{M}(\mathbf{x}) \, | \, \mathbf{x} \in \mathbf{X}\}$.}
\label{tab:mnist_rules}
\vspace{0.2em}
\begin{tabular}{l l p{0.55\linewidth}}
\toprule
\textbf{Rule} & \textbf{Label definition} & \textbf{Interpretation} \\
\midrule
$\texttt{4} \land \texttt{7}$ 
& $y = \bm{1}_{\{\texttt{4} \in \mathcal{M}(\mathbf{X}) \,\wedge\, \texttt{7} \in \mathcal{M}(\mathbf{X})\}}$
& Positive if both digit \texttt{4} and digit \texttt{7} occur in the bag. \\[0.5em]

$\texttt{4} \lor \texttt{7}$ 
& $y = \bm{1}_{\{\texttt{4} \in \mathcal{M}(\mathbf{X}) \,\vee\, \texttt{7} \in \mathcal{M}(\mathbf{X})\}}$
& Positive if digit \texttt{4}, digit \texttt{7}, or both occur in the bag. \\[0.5em]

$\texttt{4} \rightarrow \texttt{7}$ 
& $y = \bm{1}_{\{\texttt{4} \notin \mathcal{M}(\mathbf{X}) \,\vee\, \texttt{7} \in \mathcal{M}(\mathbf{X})\}}$
& Positive if digit \texttt{4} is absent or digit \texttt{7} is present. \\
\bottomrule
\end{tabular}
\vspace{-0.8em}
\end{table}
After training three end-to-end MIL models from the simulated data, we assess whether Symb-xMIL can expose the ground-truth logical rules.

\textbf{Model training.}
We used a two-layer convolutional encoder trained jointly with TransMIL~\cite{shao2021transmil} from random initialization. All models achieved near-perfect test performance (99.99\%), confirming that the simulated interaction patterns were learned by the corresponding model. Details can be found in Appendix~\ref{appendix:training_details:simulation}, also providing experimental evidence on the robustness of the MIL model to bag-size variability.

\begin{table}[!b]
\caption{\textbf{Results of simulated experiments.} The top row shows the Top-3 symbolic queries aligned with the Transformer-based MIL model trained to learn the logic of each dataset. The bottom row shows the ground-truth queries corresponding to the other two synthetic datasets. The alignment score (see Eq.~\ref{eq:sybm_xmil_corr}) is averaged over all test bags with a positive label ($y=1$). Symb-xMIL identifies the ground-truth rules with the highest correlation in all cases.}
\centering
\vspace{0.2em}
\begin{tabular}{lc|lc|lc}
\multicolumn{2}{c}{Dataset: $\mathtt{4} \land \mathtt{7}$} &
\multicolumn{2}{c}{Dataset: $\mathtt{4} \lor \mathtt{7}$} &
\multicolumn{2}{c}{Dataset: $\mathtt{4} \rightarrow \mathtt{7}$} \\

\textbf{Query} & \textbf{Mean alignment} &
\textbf{Query} & \textbf{Mean alignment} &
\textbf{Query} & \textbf{Mean alignment} \\
\midrule

$\mathtt{4} \land \mathtt{7}$ & \textbf{0.951} &
$\mathtt{4} \lor \mathtt{7}$ & \textbf{0.992} &
$ \mathtt{4} \rightarrow \mathtt{7}$ & \textbf{0.967} \\

$\mathtt{4} \land (\mathtt{7} \lor \mathtt{R})$ & 0.761 &
$\mathtt{4} \lor \neg \mathtt{R}$ & 0.675 &
$\mathtt{7}$ & 0.779 \\

$\mathtt{4}$ & 0.658 &
$\mathtt{7} \lor \neg \mathtt{R}$ & 0.669 &
$\mathtt{7} \lor \mathtt{R}$ & 0.659 \\
\midrule
$\mathtt{4} \lor \mathtt{7}$ & 0.392 & $\mathtt{4} \land \mathtt{7}$ & 0.283 & $\mathtt{4} \land \mathtt{7}$ & 0.297\\
$\mathtt{4} \rightarrow \mathtt{7}$  & 0.230 & $\mathtt{4} \rightarrow \mathtt{7}$  & -0.246 & $\mathtt{4} \lor \mathtt{7}$ & -0.145\\
\end{tabular}
\label{tab:imp_qs}
\vspace{-1em}
\end{table}

\textbf{Semantic map.}
Since ``$\texttt{4}$'' and ``$\texttt{7}$'' were the main digits of our simulations, we grouped bag instances into three semantic features, $\mathcal{N}=\{\texttt{4},\texttt{7},\texttt{R}\}$, where ``$\texttt{R}$'' stands for \textit{rest}, i.e., all digits except 4 and 7.

\textbf{Symb-xMIL results.}
Following Eq.~\ref{eq:symbolic_space}, we constructed a representative query space over the semantic features of interest, i.e., $\mathcal{N}=\{\texttt{4},\texttt{7},\texttt{R}\}$. For this synthetic experiment, with $|\mathcal{N}|=3$, this yields 39 non-redundant symbolic queries, such as $\texttt{4}$, $\neg \texttt{7}$, and $\neg \texttt{4} \land \texttt{7} \land \texttt{R}$. Details of query generation are available in Appendix~\ref{appendix:query_space}.

Using the trained models, we then used Symb-xMIL for identifying the top query for each dataset (following Eq.\eqref{eq:q_star}). In all three datasets, the recovered top-ranked query matched the ground-truth simulated logic (Table~\ref{tab:imp_qs}), while achieving substantially higher average alignment score than other queries. This shows that Symb-xMIL not only recovers the correct symbolic structures in these simulations, but also clearly distinguishes them from alternative explanations.

For comparison, we also evaluated Symbolic XAI method~\cite{schnake2025towards} in the same setting. It recovers the ground-truth rule on the \texttt{4} $\rightarrow$ \texttt{7} dataset, but is more limited on the \texttt{4} $\land$ \texttt{7} and \texttt{4} $\lor$ \texttt{7} datasets, where the top-ranked query differs from the underlying logic (detailed in Appendix~\ref{appendix:symbolic_xai}). 

\subsection{Identifying hidden model errors in tumor classification}
\label{experiment_camelyon16}

To demonstrate the practical value of symbolic explanations for identifying model errors and Clever Hans strategies—where predictions are correct for the wrong reasons \cite{lapuschkin2019unmasking, kauffmann2025explainable, linhardt2024preemptively}—we apply Symb-xMIL to a real-world histopathology task.
In particular, we focus on tumor detection from whole slide images (WSIs). We used Camelyon16~\cite{bejnordi2017diagnostic}, which is a dataset of WSIs of breast lymph nodes and binary labels indicating whether they contain metastatic tumor cells ($y=1$) or not ($y=0$)--dataset details in Appendix \ref{appendix:training_details:camelyon}. We posit that a correctly functioning model should base positive predictions on instances that actually contain tumor cells.

\textbf{Model training.}
We trained a TransMIL model \cite{shao2021transmil} on top of features extracted from a Virchow2 backbone \cite{zimmermann2024virchow2} from slide patches extracted at 20x magnification, using a stratified 5-fold cross-validation (CV) approach (training details in Appendix~\ref{appendix:training_details:camelyon}). The 5 CV-selected models achieved an almost perfect mean AUROC of $0.995\!\pm\!0.010$ on the held-out test data, suggesting that they potentially learned the underlying biology.

\vspace{-1em}
\paragraph{Semantic map.}\label{camelyon16_seg_map}
In addition to slide-level tumor labels, Camelyon16 provides pixel-level annotations distinguishing \textit{Background}, \textit{Tumor}, and \textit{Normal}. The dataset contains both macro- and micrometastatic cases. We preprocessed each WSI (detailed in Appendix~\ref{appendix:training_details:preprcessing}) and derived a semantic feature set by assigning each patch $\mathbf{x}_i$ according to its tumor-pixel fraction $\rho_i$ to $\texttt{Normal}$, $\texttt{Tumor}$, and $\texttt{Mix}$ for $\rho_i=0$, $\rho_i=1$, and $0<\rho_i < 1$, respectively. This yields $\mathcal{N}=\{\texttt{Tumor},\texttt{Mix},\texttt{Normal}\}$, where $\texttt{Mix}$ includes patches containing both tumor and normal tissue. More semantic map details in Appendix~\ref{appendix:segmentation_maps:camelyon}.
\begin{figure*}[b!]
    \centering
    \vspace{0em}
    \includegraphics[width=0.7\textwidth]{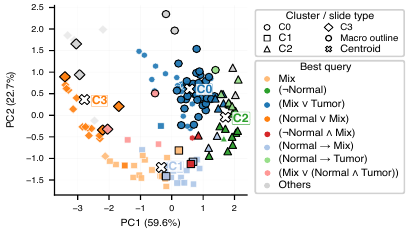}
    \caption{\textbf{PCA visualization of the Camelyon16 symbolic representation space.} Each point represents one tumor slide and is colored by its best-query group; groups with fewer than three are labeled "Others". Markers indicate cluster centers, and black outlines denote macro-metastasis slides. See also Table~\ref{tab:camelyon_clusters} for more information about the clusters.}
    \label{fig:camelyon16_pca}
    
\end{figure*}

\textbf{Symbolic representation space analysis.}
We constructed sub-bags according to the semantic map and evaluated them using the trained MIL model. Additional analysis of sub-bag embedding is provided in Appendix~\ref{sub-bag-robustess}. Following the query generation in Appendix~\ref{appendix:query_space}, the query space contains 39 logical queries for $|\mathcal{N}|=3$ (e.g., $\neg$\texttt{Normal}, \texttt{Mix} $\lor$ \texttt{Tumor}, \texttt{Normal} $\land$ (\texttt{Mix} $\lor$ \texttt{Tumor})). As a result, each of the 160 tumor-positive test slides was mapped to a point in the 39-dimensional symbolic representation space (ablation on changing query space generation in Appendix~\ref{appendix:query_space_ablation}). In order to study the model's strategy at the cohort level, we applied $K$-means clustering ($K=4$; silhouette = 0.44) to the representations of the tumor samples in the symbolic representation space.

Despite the high predictive model performance and the conceptual simplicity of the tumor detection task, the clusters reveal that there is considerable heterogeneity in model's strategy (Fig.~\ref{fig:camelyon16_pca}). Specifically, we identified four distinct symbolic decision regimes (detailed best-query statistics and example slides in Appendix~\ref{appendix:camelyon}):

    \textbf{C0} ($68$ slides), the largest cluster, is dominated by the rule \texttt{Mix $\lor$ Tumor}, meaning that the model identified sufficient evidence for the tumor-positive predictions in both the tumor areas and the mixed areas of these slides. This mirrors the expected behavior, as any patch from these two areas contains tumor cells.

    \textbf{C1} is dominated by the query \texttt{Normal $\rightarrow$ Mix}, indicating that mixed patches provide positive evidence, while normal tissue alone is insufficient and must be accompanied by mixed patches to support a positive prediction. Our investigations showed that the samples with this query are mostly samples with micrometastases without large, consistent tumor areas. 

    \textbf{C2} is dominated by macrometastasis slides (25/26 slides). The majority best query is \texttt{$\neg$Normal}, which is equivalent to \texttt{$\neg$Normal $\land$ (Mix $\lor$ Tumor)}. This indicates that the model’s tumor-positive predictions are driven by tumor or mixed areas, while the presence of normal tissue is incompatible with the decision rule and thus provides negative evidence against a positive prediction. C0 reflects a similar reliance on tumor and mixed regions, but does not enforce this exclusion of normal tissue.
    
    \textbf{C3} is characterized by the query \texttt{Normal $\lor$ Mix}, indicating that the model assigns tumor-supporting evidence not only to mixed patches containing tumor, but also to normal regions despite the absence of tumor cells. We hypothesize that these normal patches contain Clever Hans signals that spuriously drive tumor-positive predictions. To test this, we performed a counterfactual patch-transfer experiment, inserting Normal patches from C3 slides into matched ground-truth normal slides while controlling for acquisition-related confounders (Experiment details in Appendix~\ref{patch_transfer}). If these patches carry no such signal, the recipient slides should remain classified as normal. However, the counterfactual intervention flipped 21/25 cases at a 0.5 decision threshold (84\%).

To investigate whether first-order attributions could also uncover the identified Clever Hans strategy, we analyzed LRP relevance scores with respect to the semantic categories (Appendix~\ref{comparision_first_order}). However, this analysis did not reveal the same failure mode, highlighting the added value of Symb-xMIL in capturing the model's decision logic.

\subsection{Discovering sample subgroups in head-and-neck cancer}
\label{experiment_hnsc}

We explored Symb-xMIL's potential for discovering previously unknown patient subgroups. Specifically, we consider human papillomavirus (HPV) prediction directly from head-and-neck squamous cell carcinoma (HNSCC) H\&E slides. TCGA-HNSCC \cite{cancer2015comprehensive} dataset contains both HPV-positive and HPV-negative tumors, forming a binary classification task (dataset details in Appendix~\ref{appendix:training_details:hnscc}). HPV status is a clinically important biomarker that affects prognosis and treatment decisions~\cite{bilal2023aggregation}, with HPV-positive tumors generally associated with more favorable outcomes than HPV-negative cases~\cite{ang2010human}. We trained a TransMIL model \cite{shao2021transmil} on top of features extracted from a Virchow2 backbone \cite{zimmermann2024virchow2} from slide patches extracted at 20x magnification. The model achieved an overall test AUROC of $0.899\!\pm\!0.039$ (training details in Appendix~\ref{appendix:training_details:training_mil_wsi}). 

\textbf{Semantic map.}
TCGA-HNSCC dataset lacks manual semantic annotations. To obtain semantically meaningful patch labels, we constructed a two-stage labeling pipeline integrating CellViT++~\cite{horst2026cellvit++}, and tumor boundary estimation~\cite{hense2025digital} (details in Appendix~\ref{appendix:segmentation_maps:hnscc}). From the resulting spatially refined labels, we focus on the feature set
\[
\begin{aligned}
\mathcal{N} = \{&\texttt{Tumor-Neoplastic},\, \texttt{Tumor-Inflammatory},\, \texttt{Tumor-Connective}, \\
&\texttt{Non-Tumor-Inflammatory},\, \texttt{Non-Tumor-Connective}\},
\end{aligned}
\]
used for Symb-xMIL analysis. For example, semantic group \texttt{Tumor-Neoplastic} contains patches from the tumor region with predominantly neoplastic cells, while \texttt{Tumor-Inflammatory} comprises patches from the tumor region with predominantly inflammatory cells. This refinement incorporates biologically meaningful morphology and excludes implausible combinations such as \texttt{Non-Tumor-Neoplastic\footnote{Neoplastic cells are generally expected within tumor-associated regions. While isolated tumor cells may occur at invasive margins~\cite{lugli2017recommendations}, we exclude this combination to maintain consistency between tissue region and cell type annotations.}}.

\begin{figure}[!t]
    \centering
    \begin{subfigure}{\linewidth}
    \centering
    \includegraphics[width=\linewidth]{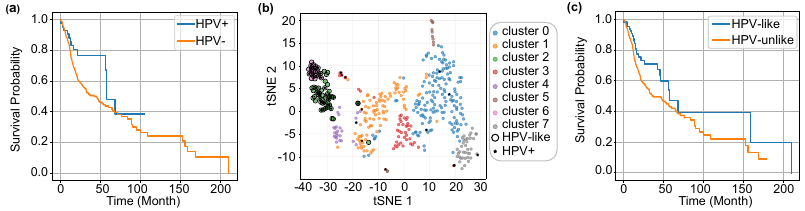}
    \end{subfigure}
    \caption{\textbf{Symb-xMIL clustering identifies HPV-like sample groups with prognostic relevance on TCGA-HNSCC.} 
    \textbf{(a)} Clinical HPV status stratifies patient survival in Kaplan-Meier analysis logrank $p=0.06$.
    \textbf{(b)} t-SNE projection of the symbolic representation shows clusters used to assign samples to HPV-like and HPV-unlike groups based on HPV+ enrichment.
    \textbf{(c)} The symbolic-space-derived HPV-like/unlike partition also stratifies survival (logrank $p=0.04$), despite being only partially concordant with clinical HPV status.}
    \label{fig:real_data_results}
    \vspace{-1em}
\end{figure}

\textbf{Survival stratification of samples in Symb-xMIL space of HPV-prediction model.} 
HPV-positive tumors are mentioned as a prognostic indicator in HNSCC \cite{ang2010human, mores2025prognostic}. This trend is also observed in TCGA-HNSCC dataset (Fig.~\ref{fig:real_data_results}-(a)), with Cox proportional hazards model (CoxPH) $z = 1.85$, one-sided $p = 0.032$ and logrank $p=0.06$. This indicates that HPV-positive patients tend to have more favorable survival outcomes, although the separation is moderate and only partially significant in this cohort.

This motivates the question of whether Symb-xMIL representations capture additional prognostic information beyond the clinical HPV status. We asked whether Symb-xMIL representations of the HPV-prediction model capture survival-relevant patterns beyond the binary clinical HPV status. To investigate this, we clustered the Symb-xMIL representations of TCGA-HNSCC samples and labeled clusters enriched for HPV+ cases as HPV-like (see Fig.~\ref{fig:real_data_results}-(b)). Here, HPV-like refers to samples that share similar model-derived morphological patterns with HPV-positive cases, independent of their clinical HPV label. This grouping allows for discordant assignments, such that some HPV+ cases are classified as HPV-unlike and some HPV- cases as HPV-like. This yields a model-driven partition of patients based on shared histomorphological patterns rather than predefined clinical labels. Methodological details are provided in Appendix~\ref{appendix:hnsc_survival}, with exemplar slides shown in Appendix~\ref{appendix:hnsc_slide}.

Based on the general positive prognostic value of HPV infection, we hypothesize that adding HPV-like samples would further refine survival association. Indeed, the symbolic HPV-like/unlike partition showed slightly stronger separation than known HPV-status with $z = 2.04$ and one-sided $p = 0.021$, logrank $p=0.04$ (Fig.~\ref{fig:real_data_results}-(c)).

A composition-matched permutation test (5,000 repetitions), preserving the same HPV+/HPV- makeup for the partitioning as the symbolic partition, showed that random assignments achieved comparable z-scores only modestly often ($\sim$9\% of the times), suggesting the observed grouping differs from random (Fig.~\ref{fig:hpv_fig_supp}-(b)). This indicates that the observed survival separation is unlikely to arise from random group assignments with the same class composition. Moreover, in an interesting observation, the HPV+ patients in the two clusters of HPV-like and HPV-unlike differed significantly in their age distributions, with the HPV-unlike ones having a significantly higher age (Fig.~\ref{fig:hpv_fig_supp}-(c)). This is consistent with the known negative prognostic effect of higher age, further supporting the clinical relevance of the identified subgroups, and demonstrating that the symbolic representation space captures implicit signals that are not directly related to the prediction target.

These results indicate that the symbolic representation space of an HPV-prediction model captures biologically meaningful patterns associated with survival, extending beyond the information contained in the binary HPV status. Because the analysis is exploratory, the findings should be perceived as hypothesis-generating. The analysis, however, showcases that symbolic-space proximity encodes HPV-like biological or morphological information that may partly recapitulate and refine the prognostic relevance of HPV-status in HNSCC cancer. 

\subsection{Ablation experiments}
In this section, we evaluate the robustness of Symb-xMIL to semantic map errors and the contribution of symbolic representations to cohort-level analysis. Our ablations show that Symb-xMIL preserves stable symbolic representations under moderate semantic map noise and reveals cohort structure less apparent from clustering raw sub-bag predictions.

\subsubsection{Robustness to semantic-map errors}
We performed sensitivity analysis of the data and symbolic explanations for the TCGA-HNSCC HPV prediction task (Section~\ref{experiment_hnsc}). We randomly selected 5--100\% of cells belonging to the \texttt{Neoplastic}, \texttt{Inflammatory}, and \texttt{Connective} classes, reassigned each selected cell uniformly to one of the other two classes, and regenerated the patch labels using the original pipeline while retaining the tumor-region information. We assessed the sensitivity of our pipeline in three axes: (1) how the patch labels change under these corruptions, (2) how the resulting symbolic representations change, and (3) how the best-aligned queries change.

As shown in Table~\ref{tab:noise-propagation}, at 50\% corruption, an average of 19.86\% of patch labels changed, while the symbolic representations remain stable, with a mean representation similarity of 0.96. The best-aligned query changed for 20.51\% of slides, indicating greater sensitivity than the full symbolic representation. This is expected since the symbolic representation space has richer information among queries and multiple queries may be correlated. 

\begin{table}[t]
    \centering
    \caption{Sensitivity to cell-label corruption on TCGA-HNSCC. We report patch-label change rates and Symb-xMIL cosine similarity between original and perturbed symbolic representations (mean and standard deviation across slides), together with the percentage of slides whose best-aligned query changes. The query space is fixed across noise levels.}
    \label{tab:noise-propagation}
    \small
    \setlength{\tabcolsep}{3pt}
    \resizebox{\linewidth}{!}{%
    \begin{tabular}{l*{11}{c}}
        \toprule
        Cell-level noise rate (\% of cells in a WSI) & 5\% & 10\% & 20\% & 30\% & 40\% & 50\% & 60\% & 70\% & 80\% & 90\% & 100\% \\
        \midrule
        Patch labels changed (\%) & 1.58 & 2.61 & 4.67 & 7.27 & 11.44 & 19.86 & 40.01 & 72.58 & 88.04 & 92.26 & 93.68 \\
        \quad Standard deviation (std, \%) & 0.61 & 0.96 & 1.64 & 2.42 & 3.43 & 4.69 & 4.84 & 5.23 & 6.40  & 6.16 & 5.92 \\
        \midrule
        Symb-xMIL cosine similarity & 1.00 & 1.00 & 0.99 & 0.99 & 0.98 & 0.96 & 0.88 & 0.45 & 0.23 & 0.17 & 0.15 \\
        \quad std & 0.01 & 0.02 & 0.03 & 0.03 & 0.05 & 0.07 & 0.09 & 0.38 & 0.44 & 0.44 & 0.43 \\
        \midrule
        Best-aligned queries changed (\%) & 3.66 & 6.18 & 9.41 & 10.57 & 14.27 & 20.51 & 43.22 & 95.60 & 97.90 & 97.67 & 97.91 \\
        \bottomrule
    \end{tabular}%
    }
\end{table}

\subsubsection{Contribution of the symbolic representation space}
We repeated the Camelyon16 (Section~\ref{experiment_camelyon16}) and TCGA-HNSCC (Section~\ref{experiment_hnsc}) clustering analyses using raw sub-bag prediction vectors $(f(X_{\mathcal S}))_{\mathcal S\subseteq\mathcal N}$ instead of the symbolic representation space, while keeping the remaining pipeline unchanged.

\begin{itemize}[leftmargin=9pt]
    \item  \textbf{Camelyon16:} Symbolic representation space clustering highlighted a potential failure mode through cluster C3, dominated by $(\mathrm{Normal}\lor\mathrm{Mix})$ (Fig.~\ref{fig:camelyon16_pca}). In contrast, raw sub-bag prediction space clustering failed to isolate all slides exhibiting the identified Clever Hans pattern, assigning ten spurious slides from C3 to clusters where this pattern was not detected. Notably, eight of these ten slides exhibited prediction flips in the patch-transfer experiment.
    \item  \textbf{TCGA-HNSCC:} The raw prediction space partition yielded $z=1.67$ (one-sided $p=0.047$) and log-rank $p=0.09$, compared with $z=2.04$ (one-sided $p=0.021$) and log-rank $p=0.04$ for the symbolic-space partition. Moreover, random permutations yielded comparable or higher z-scores less frequently for the symbolic-space partition (9\%) than for the raw prediction space partition (19\%). Together, these results suggest that the symbolic-space partition may better identify patient survival subgroups.
\end{itemize}

These exploratory findings suggest that symbolic representations make failure-related and clinically relevant cohort structure more apparent than raw sub-bag prediction vectors under the chosen clustering pipeline.

\section{Conclusion}
We introduced \textbf{Symb-xMIL}, a post-hoc framework that characterizes MIL model behavior through human-readable logical rules over semantic features. By mapping model behavior to human-readable logical rules over user-defined semantic features, Symb-xMIL moves beyond visual attribution and enables explicit reasoning about how combinations of features drive predictions. These support both sample-level explanations and cohort-level comparisons of model decision strategies. Note that a post-hoc explanation neither implies that the model internally implements the rules nor establishes biological causality.

Our experiments recovered ground-truth rules in synthetic datasets, exposed a Clever-Hans strategy~\cite{lapuschkin2019unmasking} in Camelyon16, and identified an HPV-like subgroup with prognostic relevance in TCGA-HNSCC. These findings illustrate the potential of symbolic explanations for understanding model behavior and generating hypotheses in digital pathology. Further validation on larger cohorts and more complex tasks is needed to assess the framework's clinical utility.

While Symb-xMIL enables flexible evaluation of user-defined logical queries, its scalability is limited by the exponential growth of the number of semantic feature subsets when adding more features. Future work could explore more efficient subset selection strategies and investigate how domain knowledge can guide semantic feature selection.

\section*{Acknowledgements}
The results shown here are in part based upon data generated by the TCGA Research Network: \url{https://www.cancer.gov/tcga}. We used the codes from the following repository for MIL implementations: \url{https://github.com/bifold-pathomics/xMIL}. The query generation is inspired by the following repository: \url{https://github.com/FarnoushRJ/SymbolicXAI}. We used the pretrained CellViT++ model and the cell labeling pipeline provided in the official repository: \url{https://github.com/tio-ikim/cellvit-plus-plus}. For TCGA-HNSCC, tumor, non-tumor, and border segmentation was performed using the pipeline provided in the xMIL-Pathways repository: \url{https://github.com/bifold-pathomics/xMIL-Pathways}. This work was in part supported by the Federal Ministry of Research, Technology and Space of Germany (BMFTR) under Grants 01IS14013A-E, 01GQ1115, 01GQ0850, 01IS18025A, 031L0207D, 01IS18037A, and BIFOLD24B and by the Institute of Information \& Communications Technology Planning \& Evaluation (IITP) grants funded by the Korea government (MSIT) (No. 2019-0-00079, Artificial Intelligence Graduate School Program, Korea University and No. 2022-0-00984, Development of Artificial Intelligence Technology for Personalized Plug-and-Play Explanation and Verification of Explanation). Thomas Schnake is a postdoctoral fellow at the University of Toronto in the Eric and Wendy Schmidt AI in Science Postdoctoral Fellowship Program, a program of Schmidt Sciences. The authors declare no competing interests.

\section*{Authorship contribution statement}

\textbf{Yanqing Luo:} Conceptualization, Methodology, Software, Validation, Formal analysis, Investigation, Data Curation, Visualization, Project administration, Writing - Original Draft. \textbf{Julius Hense:} Conceptualization, Methodology, Visualization, Writing - Original Draft, Project administration, Supervision. \textbf{Niklas Prenißl:} Data Curation, Investigation. \textbf{Andreas Mock:} Investigation. \textbf{Klaus-Robert Müller:} Funding acquisition. \textbf{Thomas Schnake:} Conceptualization, Methodology, Visualization, Validation, Writing - Original Draft, Project administration, Supervision \textbf{Mina Jamshidi Idaji:} Conceptualization, Methodology, Software, Validation, Investigation, Formal analysis, Visualization, Writing - Original Draft, Project administration, Supervision. \textbf{All authors:} Writing – Review \& Editing.

{\small
\bibliographystyle{plainnat}  
\bibliography{main}      
}

\newpage
\appendix

\setcounter{figure}{0}
\renewcommand{\thefigure}{S\arabic{figure}}
\setcounter{table}{0}
\renewcommand{\thetable}{S\arabic{table}}

\section*{Appendix}

\section{Query Space}
\label{appendix:query_space}
We construct the representative query space $\mathcal Q$ by enumerating sufficiently expressive logical queries over up to $k=3$ features in $\mathcal N$, using the connectives $\land$, $\lor$, and $\neg$. Queries that are logically redundant under the exhaustive partition structure are pruned (e.g., $\neg a \land \neg b \equiv c$ for $\mathcal N=\{a,b,c\}$). 

For $|\mathcal N|=3$, this yields $|\mathcal Q|=39$ non-redundant queries. For general $|\mathcal N|$, the unpruned query space contains
\[
|\mathcal Q|
=
2|\mathcal N| + 7\binom{|\mathcal N|}{2} + 12\binom{|\mathcal N|}{3}.
\]
Table~\ref{tab:query_examples} summarizes representative query classes and their semantic interpretations.

\begin{table}[htbp]
\centering
\caption{Representative categories of symbolic queries in the query space for three features. The table illustrates atomic, negated, conjunctive, disjunctive, and higher-order mixed logical forms used to construct the non-redundant symbolic query set.}
\begin{tabularx}{\linewidth}{l X X}
\toprule
\textbf{Category} & \textbf{Example} & \textbf{Interpretation} \\
\midrule
Atomic        
& $a$                
& contains feature $a$ \\
\midrule
Negation      
& $\neg a$           
& absence of feature $a$ \\
\midrule
AND           
& $a \land b$        
& contains both features $a$ and $b$ \\
\midrule
OR            
& $a \lor b$         
& contains at least one of $a,b$ \\
\midrule
AND + neg.    
& $a \land \neg b$, $\neg a \land b$
& \makecell[l]{conjunction with exclusion: pre-\\sence of one
feature and absence\\ of another} \\
\midrule
OR + neg.     
& $\neg a \lor b$, $a \lor \neg b$, $\neg a \lor \neg b$
& \makecell[l]{disjunctive rules involving ab-\\sence, including
implication-type\\ relations} \\
\midrule
Ternary AND/OR   
& $a \land b \land c$, $a \lor b \lor c$
& \makecell[l]{all-feature conjunction or at-\\least-one disjunction} \\
\midrule
Ternary AND/OR + neg. 
& $a \land b \land \neg c$, $a \land \neg b \land c$, $\neg a \land b \land c$
& \makecell[l]{conjunctions over two features\\
with exclusion of a third} \\
\midrule
Ternary mixed 
& \makecell[l]{
$a \land (b \lor c)$, $b \lor (a \land c)$, \\$b \land (a \lor c)$,
$c \lor (a \land b)$, $c \land (a \lor b)$
}
& \makecell[l]{mixed higher-order queries comb-\\ining
conjunction and disjunction} \\
\midrule
Majority      
& $(a\!\land\!b)\lor(a\!\land\!c)\lor(b\!\land\!c)$ 
& at least two of three features present \\
\bottomrule
\end{tabularx}
\label{tab:query_examples}
\end{table}

\paragraph{Query-space design.}
The query space reflects a user-controlled trade-off between interpretability and descriptive fidelity. Each query's alignment score is independent of the other evaluated queries, whereas the best-aligned query is selected from the available set. Expanding this set may reveal a better-aligned but more complex rule; restricting it may favor simpler explanations with lower alignment. Query selection should therefore reflect the user's analytical objectives and domain expertise.

\paragraph{Computational cost.}
The main computational cost arises from evaluating semantic sub-bags and scales exponentially with the number of features. For $|\mathcal N|$ features, evaluating all nonempty subsets requires up to $2^{|\mathcal N|}-1$ MIL forward passes per sample. Once these responses are cached, additional queries can be evaluated by computing their reference vectors and correlations, without further model evaluations. Thus, higher-order queries do not require additional forward passes for a fixed feature set.

Although there are $2^{2^{|\mathcal N|}}$ Boolean functions over $|\mathcal N|$ features, practical analyses need not enumerate this full space. Restricting query complexity reduces both the scoring cost and the burden of interpretation.

\paragraph{Feature selection.}
For a larger number of features, related features can be merged into coarser groups, such as combining macrophages and lymphocytes into inflammatory cells. This reduces the number of subsets at the expense of semantic detail, while alternative groupings provide different perspectives on model behavior. Our experiments demonstrate that small, curated feature sets can already reveal informative decision patterns.

Deliberately changing feature definitions also changes the scope of the explanation and should be distinguished from annotation noise, where the intended definitions remain fixed but instance labels are incorrect. Future work could examine how users with different domain backgrounds select features and queries and interpret the resulting explanations.

\subsection{Ablation: Influence of query space on symbolic representation} \label{appendix:query_space_ablation}

We evaluate the sensitivity of the symbolic representation space to the choice of query space. For this ablation, we repeat the Camelyon16 analysis using a reduced query space excluding negation operators, yielding an 18-dimensional symbolic representation space.

Clustering in the reduced query space remains highly consistent with the original clustering from Section~\ref{experiment_camelyon16}, achieving 98.5\% cluster alignment. In particular, the C0 and C3 clusters are nearly unchanged, with over 95\% agreement in slide-level best-query assignments.

For the remaining clusters, some best-query labels change due to the removal of negated expressions, but these are largely replaced by semantically related alternative queries rather than fundamentally different explanation regimes. For example, for the best query group \texttt{$\neg$Normal} in Fig.~\ref{fig:camelyon16_pca}, it becomes to \texttt{Mix $\lor$ Tumor}

Overall, these results suggest that meaningful global structure is driven by the model's subset-level behavior and that the query space is expressive enough. Although an individual best query may vary under different query space constructions, such variation primarily reflects alternative symbolic descriptions of the same explanation pattern, supporting the interpretation of the best query as a reference description rather than a uniquely fixed symbolic rule in the real-world datasets.

\section{Comparison with other explanation methods}

\subsection{Comparison with Symbolic XAI}\label{appendix:symbolic_xai}
In this section, we report additional experiments using the original Symbolic XAI method~\cite{schnake2025towards} on our synthetic MIL datasets. This experiment serves as a motivation for the modification introduced in Section~\ref{method:symb_mil}, where we compute the alignment directly from model predictions on semantic subsets instead of using Harsanyi-dividends-based~\cite{besner2020value} higher-order relevance scores.

Following Symbolic XAI, for a query $q$ and a bag $\mathbf{X}$, we compute the alignment score as
\[
r_{\mathbf{X}}(q, f)
:=
\operatorname{corr}_{\mathcal{S} \subseteq \mathcal{N}}
\left(
\lambda_q(\mathcal{S}),
\mu(\mathbf{X}_{\mathcal{S}})
\right),
\]
where $\lambda_q(\mathcal{S})$ denotes the binary reference vector induced by query $q$, and $\mu(\mathbf{X}_{\mathcal{S}})$ denotes the higher-order relevance assigned to subset $\mathcal{S}$:
\[
\mu(\mathbf{X}_{\mathcal{S}})
=
\sum_{\mathcal{L} \subseteq \mathcal{S}}
(-1)^{|\mathcal{S}|-|\mathcal{L}|}
f(\mathbf{X}_{\mathcal{L}}).
\]

As shown in Table~\ref{tab:symb_xai_synthetic}, the original Symbolic XAI method does not always recover the ground-truth rule in the MIL setting. For the $\mathtt{4} \land \mathtt{7}$ dataset, the top-ranked query is the more restrictive rule $\neg \mathtt{R} \land \mathtt{4} \land \mathtt{7}$ rather than the ground-truth rule $\mathtt{4} \land \mathtt{7}$. For the $\mathtt{4} \lor \mathtt{7}$ dataset, the top-ranked query is $\neg \mathtt{4} \lor \neg \mathtt{7}$, while the ground-truth rule $\mathtt{4} \lor \mathtt{7}$ appears only at rank four. In contrast, for the $\mathtt{4} \rightarrow \mathtt{7}$ dataset, the ground-truth rule is recovered as the top-ranked query, with a moderate alignment correlation of 0.67 compared to the proposed alignment score in this work which recovers this rule with an alignment score of 0.97 (see Table \ref{tab:imp_qs}). These results indicate that Symbolic XAI can capture meaningful symbolic structure, but its relevance-decomposition-based formulation is not always directly aligned with the bag-level decision logic learned by MIL models. This motivates our use of subset-level model predictions for computing symbolic alignment in Symb-xMIL.

\begin{table}[htbp]
\centering
\caption{\textbf{Results of simulated experiments using Symbolic XAI method.} 
Each block reports the top-5 symbolic queries ranked by their mean alignment with the Transformer-based MIL model trained on the corresponding synthetic dataset. The last row reports the alignment score of the ground-truth query when it is not included among the top-5 queries.}
\label{tab:symb_xai_synthetic}

\small
\setlength{\tabcolsep}{3pt}
\renewcommand{\arraystretch}{1.05}
\vspace{0.2em}
\begin{tabular}{lc|lc|lc}

\multicolumn{2}{c}{Dataset: $\mathtt{4} \land \mathtt{7}$} &
\multicolumn{2}{c}{Dataset: $\mathtt{4} \lor \mathtt{7}$} &
\multicolumn{2}{c}{Dataset: $\mathtt{4} \rightarrow \mathtt{7}$} \\

\textbf{Query} & \textbf{Mean alignment} &
\textbf{Query} & \textbf{Mean alignment} &
\textbf{Query} & \textbf{Mean alignment} \\
\midrule

$\neg \mathtt{R} \land \mathtt{4} \land \mathtt{7}$ & 0.730 &
$\neg \mathtt{4} \lor \neg \mathtt{7}$ & 0.713 &
$\mathtt{4} \rightarrow \mathtt{7}$ & \textbf{0.666} \\

\makecell[l]{$(\mathtt{R} \land \mathtt{4}) \lor (\mathtt{R} \land \mathtt{7})$\\
$\lor\,(\mathtt{4} \land \mathtt{7})$} & 0.674 &
$\mathtt{4} \land \neg \mathtt{7}$ & 0.529 &
$\neg \mathtt{R} \land \mathtt{7}$ & 0.618 \\

$\mathtt{4} \lor (\mathtt{R} \land \mathtt{7})$ & 0.545 &
$\neg \mathtt{4} \land \mathtt{7}$ & 0.525 &
$\mathtt{R} \lor \mathtt{7}$ & 0.491 \\

$\mathtt{7} \lor (\mathtt{R} \land \mathtt{4})$ & 0.518 &
$\mathtt{4} \lor \mathtt{7}$ & \textbf{0.440} &
$\mathtt{7}$ & 0.378 \\

$\mathtt{4} \land (\mathtt{R} \lor \mathtt{7})$ & 0.507 &
$\mathtt{R} \land \mathtt{4} \land \neg \mathtt{7}$ & 0.257 &
$\mathtt{4} \land \mathtt{7}$ & 0.340 \\
\midrule
$\mathtt{4} \land \mathtt{7}$ & \textbf{0.301} &
-- & -- &
-- & -- \\
\end{tabular}
\end{table}

\subsection{Comparison to first-order heatmaps}
\label{comparision_first_order}
To assess whether feature-level summaries of first-order attributions reveal the C3 failure mode in Camelyon16 (Section~\ref{experiment_camelyon16}), we used xMIL to get the LRP value for each patch, and we summed LRP relevance by semantic class. Among the 21 slides whose \texttt{Normal} patches induced prediction flips in the patch-transfer experiment, only 10 showed positive aggregate LRP relevance for \texttt{Normal}.
We also computed area-normalized absolute-relevance enrichment for the 25 slides in cluster C3 (Fig.~\ref{fig:camelyon16_pca}), since summed relevance may increase simply because more patches are assigned to a feature. This metric compares the fraction of total absolute relevance assigned to a feature with the fraction of patches belonging to that feature: \[
\operatorname{Enrichment}_{\mathcal S}
=
\frac{|I|\sum_{i\in I_{\mathcal S}}|r(i)|}
     {|I_{\mathcal S}|\sum_{i\in I}|r(i)|}.
\] where $r(i)$ denotes the LRP relevance of patch $i$, $I$ is the set of all patch indices in the slide, and $I_{\mathcal S}
=\{i\in I:\mathcal M(\mathbf x_i)\in\mathcal S\}$ contains the indices of patches assigned to features in $\mathcal S$. For feature-wise enrichment, $\mathcal S$ contains a single feature. The \texttt{Normal} enrichment score was below one in all slides (median 0.44, range 0.242--1.00), whereas \texttt{Mix}(median 91.6, range 3.72–1609.35) was consistently enriched.

Thus, neither summed feature-level relevance nor absolute-relevance enrichment consistently revealed the experimentally demonstrated role of \texttt{Normal} tissue in inducing tumor-positive predictions. Semantic maps alone describe tissue composition; aggregating attributions within these maps connects features to the model's predictions but does not explicitly characterize their logical relationships. Symb-xMIL goes beyond this aggregation by characterizing model responses to combinations of features. These results highlight its added value in this case, rather than a general limitation of LRP: first-order attribution and symbolic alignment capture different aspects of model behavior.

\section{Dataset and training details}
\label{appendix:training_details}

\subsection{Simulations and bag-size sensitivity}\label{appendix:training_details:simulation}
We create a dataset for each rule by sampling 2000 bags for training, 500 bags for validation, and 200 bags as held-out test samples. To promote cross-bag generalization, we train using variable bag sizes between 1 and 30 instances. Models were optimized for up to 200 epochs with early stopping and typically converged after 30--40 epochs, depending on the random seed.

\begin{figure}[htbp]
    \centering
    \includegraphics[width=\textwidth]{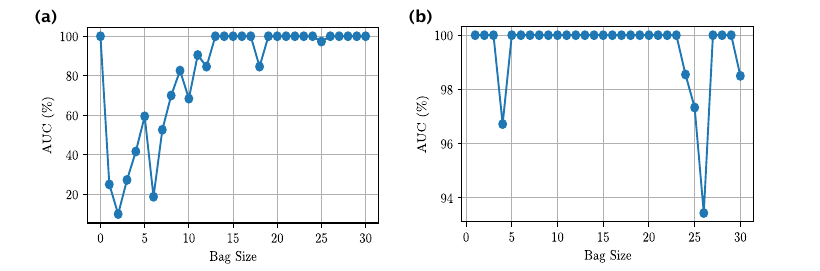}
    \caption{\textbf{Influence of variable bag size training on model performance in the simulation setting.} 
    (a) Training with a fixed bag size (30) leads to poor generalization on smaller bags. 
    (b) Perturbed training with variable bag sizes (1–30) improves robustness across different bag sizes.}
    \label{fig:auc_bag_comparison}
\end{figure}

Initially, symbolic explanations were unstable because models trained only on a fixed bag size (30) generalized poorly to perturbed bags of different sizes. As shown in Fig.~\ref{fig:auc_bag_comparison}-(a), performance degrades substantially for small bags (1--12 instances), with accuracy dropping to 15\% in the worst case. This sensitivity is problematic for Symb-xMIL, whose perturbation-based query evaluation requires robust predictions across varying bag configurations.

To address this, we introduced variable bag-size training, where bag size is randomly sampled from 1 to 30 during training. As shown in Fig.~\ref{fig:auc_bag_comparison}-(b), this yields consistently strong performance across bag sizes, with minimum accuracy above 92\%.

Additionally, for the query $4 \rightarrow 7$, an empty bag constitutes a special case: by the logical rule it evaluates to 1, whereas our default empty-bag model output is 0. We therefore treat this as an exception during evaluation to preserve consistency with the dataset logic. 

For the simulation setting, variable bag size arises naturally from the dataset construction. We also tested the same variable bag-size strategy for the Camelyon16 dataset, randomly sampling 500 to 5000 instances per bag during training. A comparison with a fixed bag size (2048) turned into similar model performance and symbolic explanation patterns. In this Camelyon16 experiment, fixed-size training yielded similar predictive performance and symbolic explanation patterns.

\subsection{Training and testing of MIL models on WSIs} \label{appendix:training_details:training_mil_wsi}
We used the checkpoints from \cite{idaji2026beyond}. 
We applied 5 folds cross-validation with five predefined splits. In each fold, the TransMIL model(without PPEG and residual part in attention) was trained on the three training splits, one split was used for validation, and the remaining split was held out for testing. 
For model selection, we conducted a grid search over attention dropout $\in \{0, 0.5\}$, classifier dropout $\in \{0, 0.5\}$, feature dropout $\in \{0, 0.2\}$, learning rate $\in \{2\times10^{-5}, 2\times10^{-4}, 2\times10^{-3}\}$, weight decay $\in \{0, 0.01\}$, and gradient clipping. The bag size and batch size were fixed to 2048 and 5, respectively, and we randomly sampled 2048 patches from each training bag. This random selection could improve the robustness to changes in both bag size and composition. All models were trained for 200 epochs and were optimized with SGD using a task-specific loss function. For each cross-validation fold, we selected the final hyperparameter configuration based on validation-set performance.

Throughout the manuscript, for each slide, the reported/used prediction was taken from the fold in which that slide belonged in the held-out test set.

\paragraph{Compute resources.}
All experiments were conducted on an internal computing cluster using NVIDIA GPU workers. MIL model training was performed on GPU workers; in our experiments, a 5-hour wall-time limit per training job was sufficient. Symb-xMIL evaluation, including query generation and alignment computation, was performed on CPU workers using eight cores. 

For Camelyon16 (399 slides, three features, seven nonempty subsets and 39 queries per slide), evaluation took a mean wall-clock time of 1.62 min per fold, with a mean total CPU time of 6 min 44 s and peak memory usage of 5.70 GiB. For TCGA-HNSCC (433 slides, five features, 31 nonempty subsets and 200 queries per slide), the corresponding values were 6.55 min, 34 min 55 s, and 7.01 GiB.

\subsubsection{Preprocessing of WSIs} \label{appendix:training_details:preprcessing}
We applied Otsu-based tissue segmentation to exclude non-informative background regions and extract \(256 \times 256\) pixel patches from tissue areas. Then we pass these extracted patches to a Virchow2~\cite{zimmermann2024virchow2} model to get the 1280-dimensional feature vectors. Our preprocessing pipeline is similar to that of \cite{hense2024xmil, hense2025digital, idaji2026beyond}.

\subsubsection{Tumor classification on Camelyon16 dataset}\label{appendix:training_details:camelyon}
The dataset comprises 399 WSIs, including 239 normal and 160 tumor slides. We trained a TransMIL classifier using cross-entropy loss for tumor detection using 5-fold cross-validation, preserving the tumor-to-normal ratio across folds. This results in test folds containing approximately \(47.8 \pm 1.9\) normal and \(32.0 \pm 1.9\) tumor slides, indicating stable class composition across splits.

\subsubsection{Patch-transfer experiment on Camelyon16}
\label{patch_transfer}
For each slide in cluster \textbf{C3} (Fig.~\ref{fig:camelyon16_pca}), we selected a tumor-free recipient slide from the same acquisition site and held-out fold to control for site-related acquisition differences and ensure evaluation with the same trained checkpoint. All recipient slides were classified as tumor-free before transfer.

We replaced randomly selected recipient patches with \texttt{Normal} patches from the C3 donor slide, preserving the recipient bag size. If the donor contained fewer \texttt{Normal} patches than the recipient's total patch count, all donor \texttt{Normal} patches were transferred. Otherwise, a random subset of donor \texttt{Normal} patches was selected to replace all recipient patches. Predictions were compared before and after transfer at a decision threshold of $0.5$, yielding negative-to-positive flips in 21 of the 25 pairs (84\%).

\subsubsection{HPV status classification on TCGA-HNSCC dataset}\label{appendix:training_details:hnscc}
We trained a TransMIL model for slide-level HPV status classification (HPV Positive $y=1$ and HPV Negative $y=0$). The dataset comprises 433 slides, including 390 HPV-negative and 43 HPV-positive samples, resulting in a pronounced class imbalance. We employ stratified 5-fold cross-validation with cross-entropy loss to preserve this distribution across splits, yielding test folds with approximately \(78.0 \pm 3.7\) HPV-negative and \(8.6 \pm 2.8\) HPV-positive slides.

\subsection{Robustness to sub-bag perturbations on Camelyon16 dataset}
\label{sub-bag-robustess}

Semantic sub-bag construction is intended to change the model input and thereby probe how its predictions respond to the presented features. These response changes are precisely the signal summarized by the alignment scores. The relevant concern is therefore whether the observed responses are dominated by out-of-distribution effects introduced by the sub-bag construction. To assess this, we conducted an additional analysis in the model's representation space on Camelyon16.

For every sub-bag $\mathbf{X}_\mathcal{S}$, we extracted the slide-level representation $h(\mathbf{X}_\mathcal{S})\in\mathbb{R}^{256}$ from the layer preceding the classification head of the MIL model, and measured its mean Euclidean distance to the $k=5$ nearest representations of real, unperturbed slides. All sub-bag and reference-slide representations used in each distance comparison were extracted using the same trained checkpoint. The reference set comprises all 399 evaluated Camelyon16 slides, both tumor slides (160) and tumor-free slides (239). 

To characterize distances typically observed within the reference set, we applied the same measure to the real, unperturbed slides: For each slide, we compute the mean distance to its five nearest neighbors among the remaining slides, excluding the slide itself. We define the central 5th--95th percentile range of this leave-one-out distance distribution as the empirical reference interval. A sub-bag representation falling within this interval has a nearest-neighbor distance comparable to that among whole slide images.

We excluded cases in which removing a feature leaves the bag unchanged because that feature is not present in the slide. For example, a $\{\texttt{Mix},\texttt{Tumor}\}$ sub-bag is excluded when the original slide contains no \texttt{Normal} patches. After these exclusions, the analysis contains 872 sub-bags.

Across the 872 evaluated sub-bags, 625 ($71.7\%$) fall within the empirical reference interval. In Table \ref{tab:ood}, we additionally report, for each feature subset, the mean sub-bag size, the mean distance to the reference set (5 nearest neighbors), and the within-interval proportion of sub-bags.

Sub-bags from which the \texttt{Normal} feature is removed exhibit larger representation shifts. However, this shift does not lead to unstable prediction directions: all evaluated sub-bags without \texttt{Normal} patches retain positive model outputs. These results show that most evaluated sub-bags have nearest-neighbor distances comparable to those of unperturbed slides, while removing \texttt{Normal} tissue induces larger representation shifts without changing the tumor-positive classifications.

\begin{table}[htbp]
    \centering
    \small
    \caption{Overview of the sub-bags and their distances to the reference set. For each feature subset, \textit{mean bag size} denotes the average number of patches across all corresponding sub-bags. \textit{Mean distance} denotes the average Euclidean distance from each sub-bag representation to its five nearest real, unperturbed slide representations in the reference set. The final column reports the proportion of sub-bags whose distances fall within the central  5th--95th percentile range of the leave-one-out distances observed within the reference set.}
    \begin{tabular}{lrrr}
        \toprule
        Feature Subset  & Mean bag size & Mean dist. & Within reference interval (\%) \\
        \midrule
        \multicolumn{4}{l}{\textit{Normal tissue retained}} \\
        \texttt{Normal, Mix} & 9019 & 1.93 & 78.4\% \\
        \texttt{Normal, Tumor} & 9464 & 2.36 & 81.9\% \\
        \texttt{Normal}   & 8758 & 3.27 & 76.9\% \\
        \cmidrule(lr){1-4}
        \multicolumn{4}{l}{\textit{Normal tissue removed}} \\
        \texttt{Mix, Tumor} &  992 & 3.62 & 66.9\% \\
        \texttt{Mix} &  286 & 3.57 & 66.2\% \\
        \texttt{Tumor}  &  974 & 4.01 & 57.8\% \\
        \bottomrule
    \end{tabular}
    \label{tab:ood}
\end{table}

\section{Semantic mapping of real-world datasets}
\label{appendix:segmentation_maps}
\subsection{Camelyon16 dataset}\label{appendix:segmentation_maps:camelyon}
\begin{figure}[htbp]
    \centering
    \includegraphics[width=\linewidth]{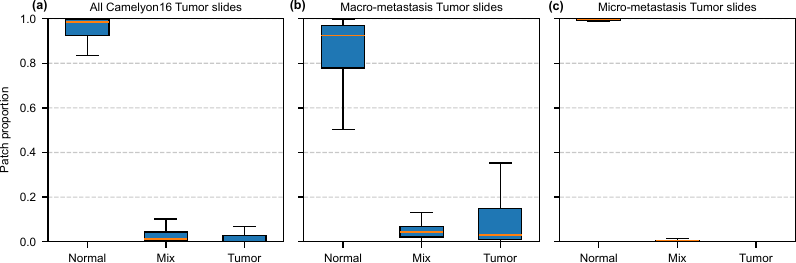}
    \caption{\textit{Distribution of semantic patch proportions in Camelyon16.} (a) proportions across all tumor slides ($N=160$). (b) proportions restricted to macro-metastatic slides ($N=80$), showing increased \texttt{Tumor} and \texttt{Mix} relative to the full tumor cohort. (c) proportion of micro-metastatic slides ($N=80$), showing large \texttt{Normal} proportion with rare \texttt{Tumor} patches.}
    \label{fig:camelyon16_patch_analysis}
\end{figure}

Using the segmentation map described in Section~\ref{camelyon16_seg_map}, we first characterize the dataset-level distribution of semantic patch labels (\texttt{Tumor}, \texttt{Mix}, and \texttt{Normal}). Figure~\ref{fig:camelyon16_patch_analysis} provides an overview of the proportion of each patch type across slides. Overall, \texttt{Normal} patches dominate the composition of most slides, reflecting the strong class imbalance at the patch level even in tumor-positive WSIs.

We further compare tissue composition between macro- and micro-metastatic slides. While both groups contain a substantial proportion of Normal patches, macro-metastatic slides exhibit systematically higher proportions of $\texttt{Tumor}$ and $\texttt{Mix}$ patches, indicating a larger tumor burden. In contrast, micro-metastatic slides tend to contain less \texttt{Tumor} and \texttt{Mix} regions. 

\subsection{TCGA-HNSCC dataset}\label{appendix:segmentation_maps:hnscc}
We first applied a pretrained CellViT++ model~\cite{horst2026cellvit++}, based on the SAM~\cite{kirillov2023segment} model, which operates at $40\times$ magnification, for cell-level segmentation and classification. Each detected cell was assigned to one of five categories: \texttt{Neoplastic}, \texttt{Connective}, \texttt{Inflammatory}, \texttt{Dead}, \texttt{Epithelial}. 

\begin{figure}[htbp]
    \centering
    \includegraphics[width=\linewidth]{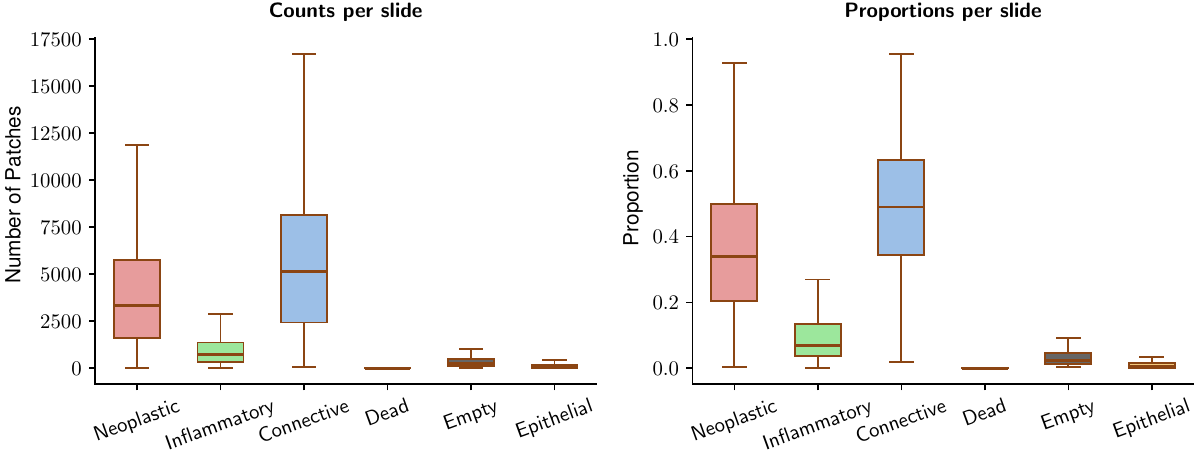}
     \caption{\textbf{Overview of patch-type counts and proportions in the TCGA-HNSCC dataset following CellViT++ segmentation and majority aggregation.} Left: boxplots showing the distribution of patch-type counts across TCGA-HNSCC slides. Right: proportions of the resulting patch types in the dataset.}   
    \label{fig:hnsc_cellvit}
\end{figure}

An analysis of patch-label distributions across the TCGA-HNSCC cohort showed that \textit{Neoplastic} and \textit{Connective} patches dominate the tissue composition of many WSIs, while \textit{Inflammatory} patches exhibit moderate but consistent presence, as shown in Fig.~\ref{fig:hnsc_cellvit}. In contrast, \textit{Dead} and \textit{Epithelial} patches were rare and often absent, whereas \textit{Empty} patches corresponded to background regions without detected cells and were excluded from further analysis. Consequently, the symbolic explanation experiments are restricted to the reduced semantic feature set consisting of \textit{Neoplastic}, \textit{Connective}, and \textit{Inflammatory} tissue types.

Second, we refined these cell-type labels by incorporating spatial context through tumor boundary estimation following~\citet{hense2025digital}. This allows each patch to be additionally classified as residing inside or outside the tumor region, yielding context-aware semantic features used for symbolic explanations.

\begin{figure}
    \centering
    \includegraphics[width=\linewidth]{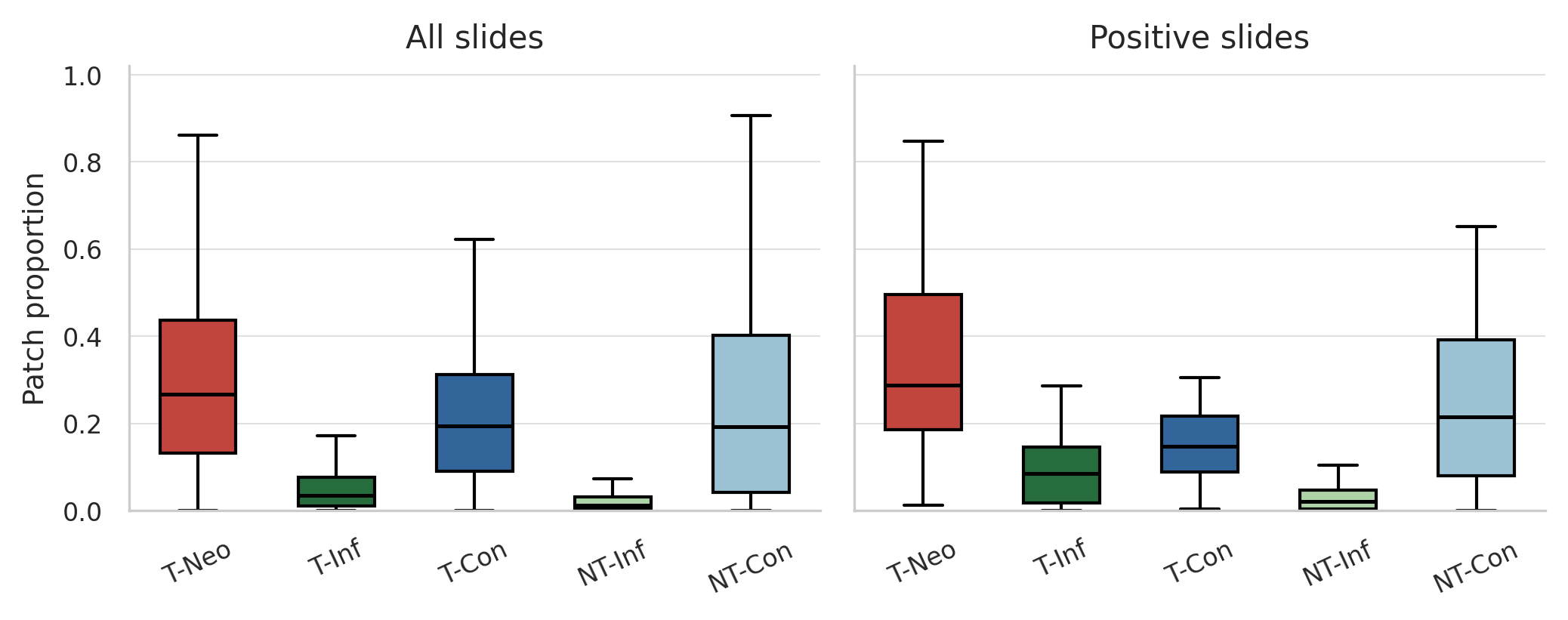}
    \caption{\textbf{Distribution of semantic patch types in TCGA-HNSCC.} Left: proportions across the full cohort. Right: proportions for HPV-positive slides. Abbreviations: T = Tumor, NT = Non-Tumor, Neo = Neoplastic, Con = Connective, Inf = Inflammatory.}
    \label{fig:hnsc_patch_analysis}
\end{figure}

Figure~\ref{fig:hnsc_patch_analysis} summarizes the resulting patch-type distributions. Across the cohort, \texttt{Tumor-Neoplastic} is the dominant feature type. Comparing HPV subgroups suggests compositional differences in the available semantic evidence: HPV-positive slides show a tendency toward higher proportions of \texttt{Tumor-Inflammatory} and \texttt{Non-Tumor-Inflammatory} patches, whereas HPV-negative slides exhibit relatively more \texttt{Tumor-Connective} patches. 

\section{Additional Analysis for the Camelyon16 Dataset}
\label{appendix:camelyon}

In this section, we provide additional analysis for Camelyon16. First, we present a detailed cluster characterization and best-query annotations for global symbolic representation space analysis shown in Section~\ref{experiment_camelyon16}. Second, we provide a detailed slide-level analysis of representative slides from each cluster, including thumbnails, semantic maps, and relevance heatmaps generated using \citet{hense2024xmil}.

\subsection{Additional Global Analysis for Camelyon16}

To further analyze structure in the symbolic representation space, we visualize the resulting clusters in a PCA projection, with each slide annotated by its best-query. The clustering does not only reflect separation between macro- and micro-metastatic cases, but also groups slides with similar symbolic rules. This suggests that the symbolic representation space captures structure related to shared decision patterns beyond metastasis subtype. Cluster-level characteristics are summarized in Table~\ref{tab:camelyon_clusters}. While some clusters are dominated by a single symbolic rule, others exhibit a mixture of related rules sharing common semantic themes. 

\begin{table}[!htbp]
    \centering
    \caption{\textbf{Characterization of symbolic explanation clusters in Camelyon16 as depicted in Fig.~\ref{fig:camelyon16_pca}.} 
For each cluster, we report the number of slides ($N$), counts of macro and micro metastasis cases, and the most frequent best-matching symbolic queries within the cluster.}
    \label{tab:camelyon_clusters}
    \small
    \begin{tabularx}{\linewidth}{@{}l c c c X@{}}
        \toprule
        \textbf{Cluster} & \textbf{N} & \textbf{Macro} & \textbf{Micro} & \textbf{Top Rules (count)} \\
        \midrule
        C0: Tumor + Context & 68 & 45 & 23 &
        \texttt{Mix $\lor$ Tumor} (63), \texttt{Mix $\lor$ (Normal $\land$ Tumor)} (1), \texttt{Normal $\lor$ Tumor} (1) \\
        C1: Context-driven & 41 & 4 & 37 &
        \texttt{$\neg$Norm $\lor$ Mix} (19), \texttt{Mix} (15), \texttt{Normal $\land$ Mix} (1) \\
        C2: Tumor-dominant & 26 & 25 & 1 &
        \texttt{$\neg$Normal} (13), \texttt{Mix $\lor$ Tumor} (4), \texttt{$\neg$Normal $\lor$ Tumor} (4) \\
        C3: Normal-driven & 25 & 6 & 19 &
        \texttt{Normal $\lor$ Mix} (16), \texttt{Normal $\land$ (Mix $\lor$ Tumor)} (2), \texttt{Normal} (2) \\
        \bottomrule
    \end{tabularx}
\end{table}

\subsection{Additional Slide-Level Analysis for Camelyon16}
\label{appendix:camelyon_slides}

In this section, we present representative slides from different clusters in Fig.~\ref{fig:camelyon16_pca} and visualize their best-aligned logical rules together with the corresponding H\&E thumbnail, semantic map, and LRP-based relevance heatmap.

Across these examples, the LRP heatmaps show relevance patterns that tend to follow the tissue patterns described by the best-aligned symbolic rules. At the same time, the example of slide \texttt{tumor\_018} highlights an important direction for future work: improving robustness to incomplete semantic maps. In this slide, two visually similar biopsy regions are present, but only one has a \texttt{Tumor}/\texttt{Mix} annotation, whereas both regions exhibit similar relevance patterns in the LRP heatmap. 

\textbf{\texttt{test\_071}, best query: \texttt{Mix $\lor$ Tumor}, cluster:C0}
 
\begin{figure}[!htbp]
\centering
\includegraphics[width=\linewidth]{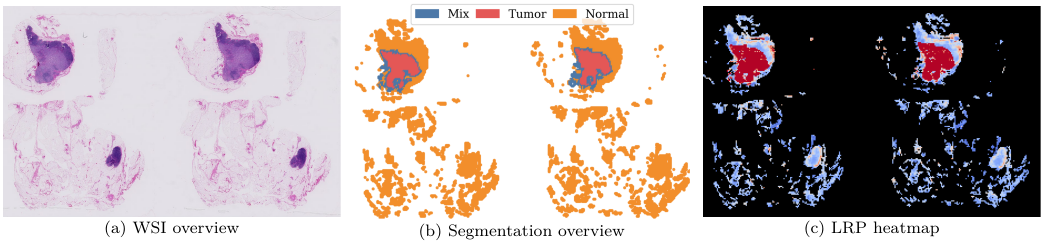}
\caption{Slide \texttt{test\_071} slide-level visualization.}
\end{figure}

\textbf{\texttt{test\_065}, best query: \texttt{Normal $\lor$ Mix}, cluster:C3}
 
\begin{figure}[H]
\centering
\includegraphics[width=\linewidth]{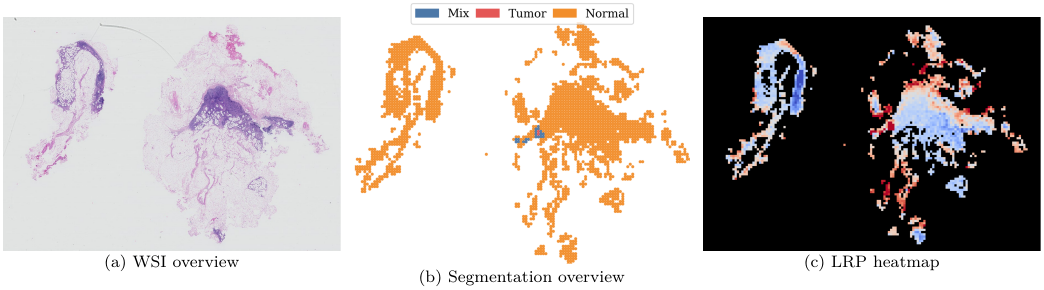}
\caption{Slide \texttt{test\_065} slide-level visualization.}
\end{figure}

\textbf{\texttt{tumor\_012}, best query: \texttt{$\neg$ Normal $\lor$ Mix}, cluster:C1}
 
\begin{figure}[H]
\centering
\includegraphics[width=\linewidth]{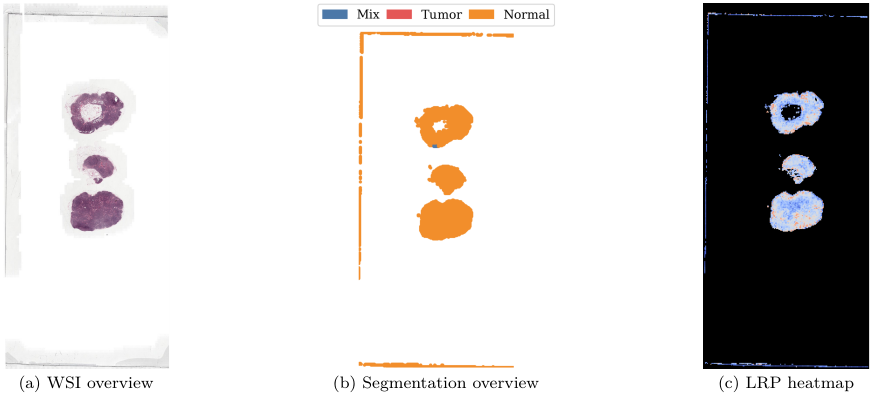}
\caption{Slide \texttt{tumor\_012} slide-level visualization.}
\end{figure}

\text{\texttt{tumor\_018}, best query: \texttt{$\neg$ Normal}, cluster:C2}
 
\begin{figure}[H]
    \centering
    \includegraphics[width=\linewidth]{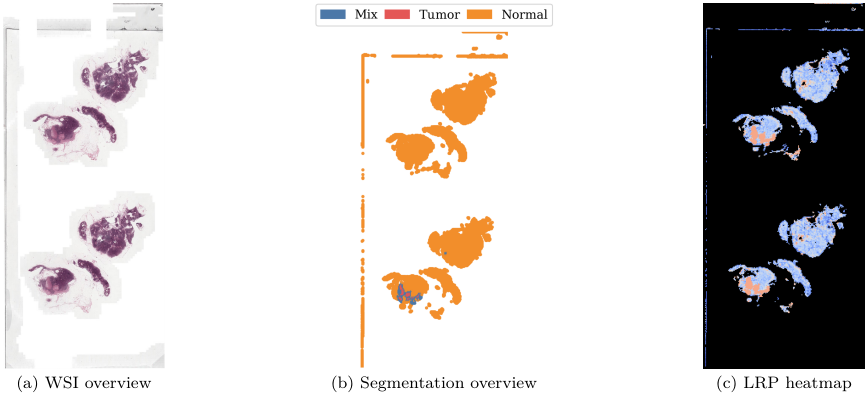}
    \caption{Slide \texttt{tumor\_018} slide-level visualization.}
\end{figure}

\section{Additional analysis for the TCGA-HNSCC Dataset}
\label{appendix:hnsc}

\subsection{Additional Details of Survival Analysis for TCGA-HNSCC}
\label{appendix:hnsc_survival}

\begin{figure}[!t]
    \centering
    \includegraphics[width=\linewidth]{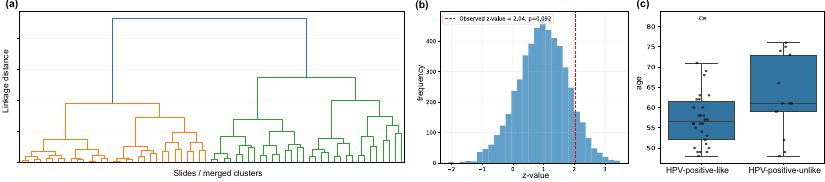}
    \caption{\textbf{Clustering of Symb-xMIL representations of TCGA-HNSCC samples computed from an HPV-prediction model.} 
    \textbf{(a)} Clustering dendrogram. \textbf{(b)} Null distribution of CoxPH z-values of randomly partitioned samples. \textbf{(c)} Patient age boxplot to HPV-positive-like and HPV-positive-unlike.}
    \label{fig:hpv_fig_supp}
\end{figure}

For clustering the symbolic representations, we applied agglomerative clustering to the standardized feature vectors of all samples in the dataset. The number of clusters was selected based on inspection of the clustering dendrogram (Fig.~\ref{fig:hpv_fig_supp}), and we used eight clusters to capture local structure in the symbolic representation space. As a robustness check, we also evaluated six- and seven-cluster solutions, which did not change the resulting HPV-like versus HPV-unlike cluster labeling. Clusters were then annotated according to their HPV+ enrichment: clusters containing a sufficiently high fraction of HPV+ samples (>20\%) were labeled as HPV-like, whereas the remaining clusters were labeled as HPV-unlike. This procedure allowed samples to be assigned based on their symbolic-space neighborhood rather than strictly by clinical HPV status, resulting in a partition that could include HPV- samples in HPV-like clusters and HPV+ samples in HPV-unlike clusters. As a result of the clustering, 13 (out of 43) HPV+ samples were grouped with HPV-unlike samples, whereas 35 (out of 390) HPV- samples were grouped with HPV-like samples. One HPV-negative sample with missing survival time was excluded from survival analysis.

We used a composition-matched permutation test to determine whether the symbolic HPV-like/unlike survival grouping of the samples resulted from clustering of Symb-xMIL representations was significantly different from random partitioning of the samples. For each permutation, we randomly selected without replacement the same number of HPV+ and HPV- samples as present in the observed HPV-like group, assigned these samples to a randomized HPV-like group, and assigned all remaining samples to HPV-unlike. We then fitted a CoxPH model to each randomized partition and recorded the resulting z-score. Fig.~\ref{fig:hpv_fig_supp} shows the resulting null distribution of z-scores. The empirical permutation p-value was calculated as the proportion of randomized partitions with z-scores greater than or equal to the observed symbolic-partition z-score, providing a null distribution that preserves HPV composition while randomizing group memberships. 

Additionally, note that studies show that there are multiple other prognostic indicators that may act as a confounder for the prognostic values of HPV-positivity in oropharyngeal cancer \cite{cadoni2017prognostic, mehanna2010head, mores2025prognostic}. Therefore, we emphasize that the presented results in this part should be taken only as a proof-of-concept to show the potential of Symb-xMIL representations, rather than clinically actionable claims.

Note that we use \textit{lifelines} \cite{Davidson-Pilon2019} for survival analyses.

\subsection{Additional Slide-Level Analysis for TCGA-HNSCC}
\label{appendix:hnsc_slide}
In this section, we provide representative slide-level visualizations for TCGA-HNSCC cases selected according to their clinical HPV status and the HPV-like/HPV-unlike grouping defined in Section~\ref{experiment_hnsc}. These examples are intended as qualitative visual checks of the learned explanation patterns. For each slide, we show the WSI overview, the semantic segmentation map, and the LRP heatmap using \citet{hense2024xmil,idaji2026beyond}.

\begin{figure}[H]
\noindent
    \textbf{\texttt{TCGA-BB-4228-01Z-00-DX1}:
    labeled as HPV+ and belongs to HPV-like class}
    \par\vspace{0.5em}
    \centering
    \includegraphics[width=\linewidth]{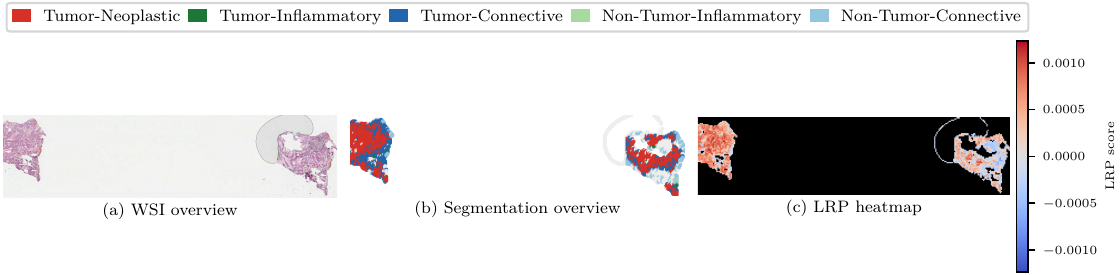}
    \caption{Slide \texttt{TCGA-BB-4228-01Z-00-DX1} slide-level visualization.}
\end{figure}

\textbf{\texttt{TCGA-HD-7832-01Z-00-DX1}: labeled as HPV+ and belongs to HPV-like class}
\begin{figure}[H]
    \centering
    \includegraphics[width=\linewidth]{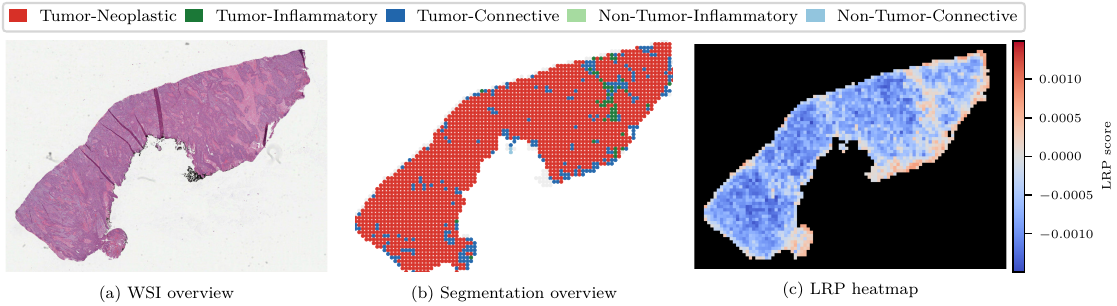}
    \caption{Slide \texttt{TCGA-HD-7832-01Z-00-DX1} slide-level visualization.}
\end{figure}

\textbf{\texttt{TCGA-CN-4741-01Z-00-DX1}: labeled as HPV+ and belongs to HPV-like class}
\begin{figure}[H]
    \centering
    \includegraphics[width=\linewidth]{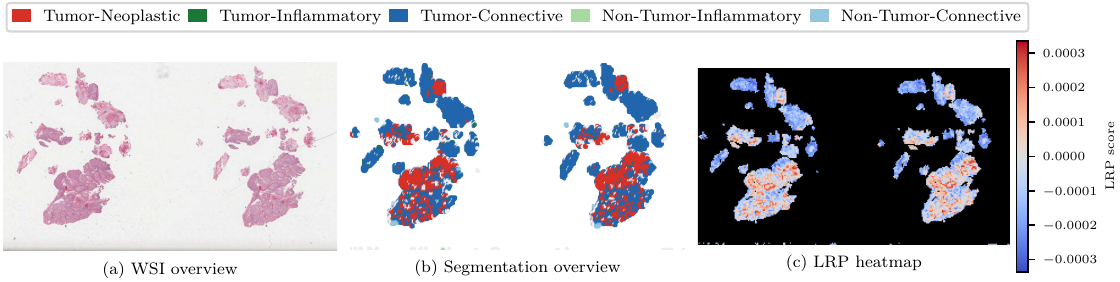}
    \caption{Slide \texttt{TCGA-CN-4741-01Z-00-DX1} slide-level visualization.}
\end{figure}

\textbf{\texttt{TCGA-CN-5360-01Z-00-DX1}: labeled as HPV- and belongs to HPV-like class}
\begin{figure}[H]
    \centering
    \includegraphics[width=\linewidth]{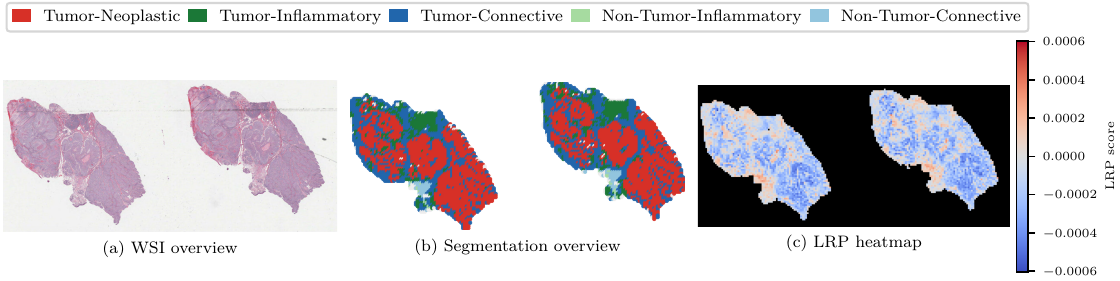}
    \caption{Slide \texttt{TCGA-CN-5360-01Z-00-DX1} slide-level visualization.}
\end{figure}

\begin{figure}[H]
\noindent
\textbf{\texttt{TCGA-BB-4225-01Z-00-DX1}:labeled as HPV+ and belongs to HPV-unlike class}
\par\vspace{0.5em}
    \centering
    \includegraphics[width=\linewidth]{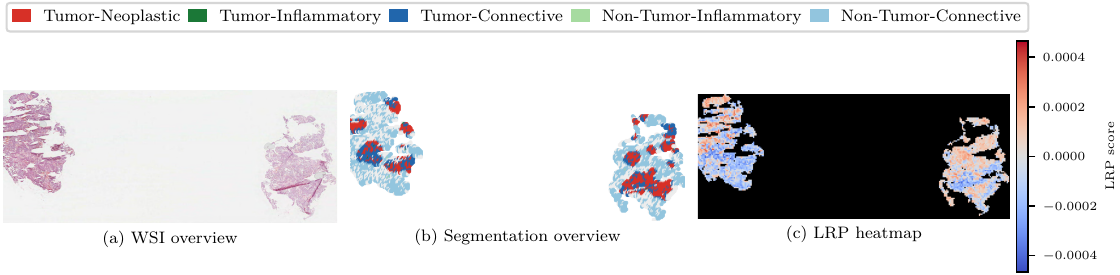}
    \caption{Slide \texttt{TCGA-BB-4225-01Z-00-DX1} slide-level visualization.}
\end{figure}

\textbf{\texttt{TCGA-T2-A6WZ-01Z-00-DX1}: labeled as HPV- and belongs to HPV-unlike class}
\begin{figure}[H]
    \centering
    \includegraphics[width=\linewidth]{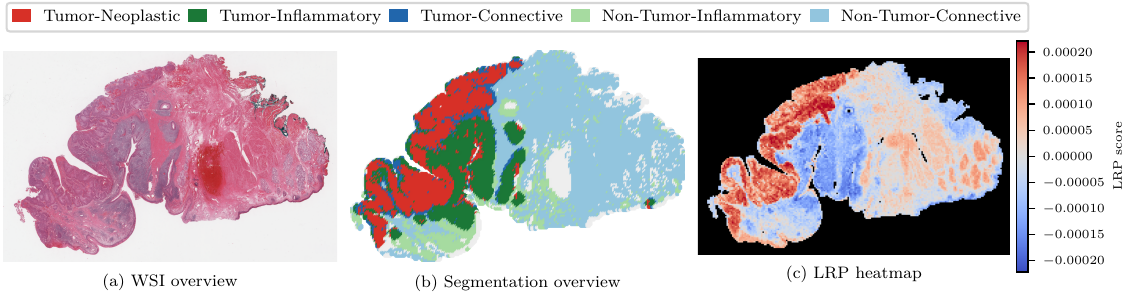}
    \caption{Slide \texttt{TCGA-T2-A6WZ-01Z-00-DX1} slide-level visualization.}
\end{figure}

\end{document}